\documentclass{article}
\usepackage[final, main, nonatbib]{neurips_2026}
\usepackage[numbers,compress]{natbib}
\usepackage[utf8]{inputenc}
\usepackage[T1]{fontenc}
\usepackage{hyperref}
\usepackage{url}
\usepackage{booktabs}
\usepackage{amsfonts}
\usepackage{nicefrac}
\usepackage{microtype}
\usepackage{xcolor}
\usepackage{todonotes}
\usepackage{graphicx}
\usepackage{subcaption}
\usepackage{amsmath}
\usepackage{float}
\usepackage[ruled,vlined]{algorithm2e}
\SetEndCharOfAlgoLine{}
\usepackage{tikz}
\usepackage{caption}
\usetikzlibrary{arrows.meta, fit, positioning, decorations.pathreplacing}
\usepackage{wrapfig}
\usepackage{booktabs}
\usepackage{colortbl}
\usepackage{array}
\usepackage{cleveref}
\usepackage{needspace}

\definecolor{first}{RGB}{102,194,165}
\definecolor{second}{RGB}{140,209,182}
\definecolor{third}{RGB}{178,224,199}
\definecolor{fourth}{RGB}{216,239,216}
\definecolor{fifth}{RGB}{240,247,237}
\definecolor{orangeFirst}{RGB}{252, 141, 98}
\definecolor{orangeSecond}{RGB}{255, 180, 150}
\definecolor{orangeThird}{RGB}{255, 200, 175}
\definecolor{orangeFourth}{RGB}{255, 220, 200}
\definecolor{orangeFifth}{RGB}{255, 240, 225}

\definecolor{purple}{RGB}{84,39,143}

\title{Epiplexity Guided Data Selection and Generation for Out-of-Distribution Generalization}

\author{%
  Ellen Su$^{*1}$\hspace{1em}
  Andres Potapczynski$^{*1}$\hspace{1em}
  Shikai Qiu$^{1}$\hspace{1em}
  Edward Hughes$^{2}$ \\[0.4em]
  \textbf{Andrew Gordon Wilson}$^{1}$ \\[0.6em]
  \normalfont $^{1}$New York University\hspace{2em}$^{2}$Inherent \\[0.2em]
  \normalfont $^{*}$Equal contribution \hspace{1.5em} 
}
\begin{document}

\maketitle

\begin{abstract}
Modern systems are increasingly expected to transfer across tasks not specified during training. What data facilitates generalization to unanticipated settings? One hypothesis is that data with more structural information could contain shared circuits and subprograms that may be recycled in a wider array of downstream settings. Epiplexity, a recently proposed measure of the structural information a compute-bounded learner can extract from data, offers a mechanism to reason about this relationship. As existing estimators for epiplexity only operate on data already seen during training, we show how to operationalize epiplexity as an online training signal by greedily maximizing its stepwise gain in two settings. For data selection, we fit scaling laws to the training loss curves of natural data domains to predict the expected epiplexity gain as a function of training tokens, and use this signal to adaptively determine the sampling weights over domains during training. For synthetic data generation, where no domain structure exists, we reward a generator with the change in epiplexity of a learner's synthetic training data and use REINFORCE policy gradients to guide the generator toward an epiplexity-maximizing distribution. In both cases, we find higher epiplexity predicts improved downstream performance on zero-shot and fine-tuning tasks, supporting the hypothesis that data rich in structural information yield representations that better transfer across domains.
\end{abstract}

\section{Introduction}
Of model, optimizer, and data, it is \emph{data} that is increasingly being recognized as the greatest lever for improving performance. This realization is compounded by an impending sense that we are entering a data-limited regime \citep{muennighoff2023scaling, villalobos2024position}, where multi-epoch training is necessary at scale.
Accordingly, there is a growing body of work exploring how training data interventions influence model performance. Current methods in data mixing and selection \cite{Xie2023DoReMiOD, Jiang2024AdaptiveDO, Liu2024RegMixDM, Fan2023DoGEDR, Chen2023SkillitAD, Gu2024DataSV, Gu2024TowardsOL, Lin2024Rho1NA}, curriculum and ordering \cite{bengio2009Curriculum, Kim2024StrategicDO}, and self-play \cite{Tesauro1995, Silver2018AGR, Chen2024SelfPlayFC, Yuan2024SelfRewardingLM, Liu2025SPIRALSO, Cheng2024SelfplayingAL, Zhao2025AbsoluteZR} all share the same implicit assumption that batches of data vary in how important they are to a learner, and that exploiting these differences can yield a model with better generalization. Many of these methods consider data quantity, diversity, and quality, but these notions are often vaguely defined. Moreover, models are increasingly deployed on new tasks and domains, many of which couldn't have been anticipated during training, making it even harder to reason about what properties of data ought to be desirable. In particular, how should we select data for greater \emph{generality} in out-of-distribution (OOD) settings?

A recent theoretical development provides a promising starting point to reason about this question. \citet{Finzi2026FromET} introduced \emph{epiplexity}, a measure of the structural information content in data, characterized as the information in  
the model that minimizes the description length of data under computational constraints. Repetitive sequences have low epiplexity because they are learnable but compressible, and noise has low epiplexity because the model cannot learn from it. In contrast, data which makes a model work harder, through sustained and substantial decreases in loss, has high epiplexity. Further, as opposed to classical information measures, \citet{Finzi2026FromET} establishes that epiplexity can be created by computation, depends on the ordering of data, and can exceed the information content of the data-generating process itself, making it a compelling signal for interrogating the relationship between the structural information of training data and model generalization. While \citet{Finzi2026FromET} contends that data with higher epiplexity should lead to better OOD generalization, this hypothesis is largely untested. They show that existing data selection methods such as ADO \citep{Jiang2024AdaptiveDO} can \emph{inadvertently} select for higher epiplexity data (relative to random batches), but neither develops a methodology that directly uses epiplexity for selection nor articulates the correspondence between epiplexity and downstream generalization. Relatedly, \citet{liu2026selfplay} uses epiplexity to diagnose LLM self-play and argues that sustained self-improvement requires generation pipelines that consistently increase epiplexity. However, they measure epiplexity post hoc and do not propose a way to optimize it. Thus, open questions remain on how to turn existing estimates for epiplexity into an online training objective.

In this paper, \textbf{we introduce the first approaches for optimizing epiplexity as an online training signal for both data selection and generation}. In both cases, providing an optimizable online estimator for epiplexity is nontrivial. For selection, we fit scaling laws to per-domain loss curves in order to predict the expected epiplexity gain as a function of in-domain training tokens, and we maximize this signal to choose optimal batches of training data. For generation, we first train a generator to produce epiplexity-maximizing data for a learner by defining the reward as the increase in the epiplexity of the learner’s training data and apply REINFORCE policy gradients \citep{zhang2021sample, Sutton1999PolicyGM} to optimize the generator parameters without external verifiers. Intuitively, as both trivial and noisy data have near-zero epiplexity, this objective creates data that challenges the learner while remaining predictable. We show that our proposed data selection (\texttt{EpiSelect}) and generation (\texttt{EpiGen}) algorithms both advance performance over the state of the art. Beyond our methodology, we discover a key limitation of a popular benchmark, \emph{The Pile} \citep{Gao2020ThePA}, in evaluating data selection methods for downstream generalization. Training on its largest domain alone outperforms state-of-the-art selection methods due to its domain size imbalances. We therefore propose shifting data selection evaluation to the more balanced Common Pile dataset \citep{kandpal2025commonpilev018tb}. Finally, we also test the hypothesis that higher epiplexity corresponds to better OOD generalization and find that it holds across several settings.

\setlength{\textfloatsep}{2pt}
\setlength{\belowcaptionskip}{10pt}
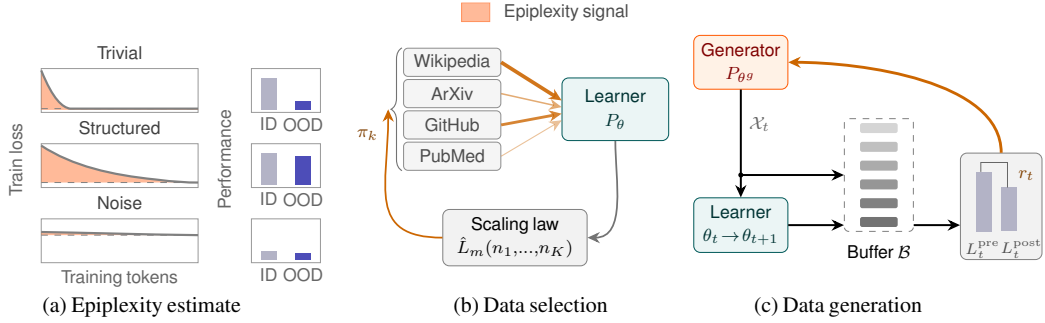
\begin{figure}[t]
\centering
\resizebox{\textwidth}{!}{
\begin{tikzpicture}[
    font=\footnotesize\sffamily,
    >={Stealth[length=2mm, width=1.8mm]},
        panel title/.style={font=\normalsize},
        dombox/.style={
        draw=black!45, fill=black!4, line width=0.3pt,
        rounded corners=2pt, minimum width=1.55cm, minimum height=0.45cm,
        inner sep=2pt, font=\footnotesize\sffamily, text=black!70
    },
    model/.style={
        draw=teal!70!black, fill=teal!8, line width=0.3pt,
        rounded corners=3pt, minimum width=1.7cm, minimum height=1.0cm,
        align=center, font=\footnotesize\sffamily, text=teal!40!black
    },
    gen/.style={
        draw=orange!70!red, fill=orange!10, line width=0.3pt,
        rounded corners=3pt, minimum width=1.7cm, minimum height=0.9cm,
        align=center, font=\footnotesize\sffamily, text=red!50!black
    },
    slaw/.style={
        draw=black!50, fill=black!5, line width=0.3pt,
        rounded corners=3pt, minimum width=2.3cm, minimum height=0.9cm,
        align=center, font=\footnotesize\sffamily
    },
    estbox/.style={
        draw=black!50, fill=black!5, line width=0.3pt,
        rounded corners=3pt,
        inner sep=4pt
    },
    buffer/.style={
        draw=black!50, line width=0.3pt, dashed,
        rounded corners=2pt, minimum width=1.7cm, minimum height=1.7cm
    },
    epi arrow/.style={->, draw=orange!80!black, line width=1.4pt},
    dom arrow/.style={->, draw=orange!80!black, line width=1.2pt},
    data arrow/.style={->, draw=black, line width=0.9pt},
    fb arrow/.style={->, draw=black!55, line width=0.8pt, dashed},
]
\begin{scope}[yshift=0cm, local bounding box=panelA]
    \node[panel title] at (1.6, -0.7) {(a) Epiplexity estimate};

    \begin{scope}[yshift=2.4cm]
        \node[font=\footnotesize\sffamily, text=black!75] at (1.25, 0.95) {Trivial};
        \draw[black!40, line width=0.3pt] (0, 0) rectangle (2.5, 0.7);
        \fill[orange!65!red, opacity=0.40]
            (0, 0.67) .. controls (0.19, 0.20) and (0.34, 0.08) .. (0.46, 0.06)
            -- (2.5, 0.06) -- (2.5, 0.04) -- (0, 0.04) -- cycle;
        \draw[black!50, dashed, line width=0.3pt] (0, 0.06) -- (2.5, 0.06);
        \draw[black!50, line width=1pt]
            (0, 0.67) .. controls (0.19, 0.20) and (0.34, 0.08) .. (0.46, 0.06)
            -- (2.5, 0.06);
        \draw[black!40, line width=0.3pt] (3.35, 0) rectangle (4.45, 0.7);
        \fill[blue!25!black!30] (3.50, 0.05) rectangle (3.75, 0.55);
        \fill[blue!60!black!70] (4.05, 0.05) rectangle (4.30, 0.18);
        \node[font=\footnotesize\sffamily, text=black!60, anchor=north] at (3.625, 0.04) {ID};
        \node[font=\footnotesize\sffamily, text=black!60, anchor=north] at (4.175, 0.04) {OOD};
    \end{scope}

    \begin{scope}[yshift=1.2cm]
        \node[font=\footnotesize\sffamily, text=black!75] at (1.25, 0.95) {Structured};
        \draw[black!40, line width=0.3pt] (0, 0) rectangle (2.5, 0.7);
        \fill[orange!65!red, opacity=0.40]
            (0, 0.67) .. controls (0.5, 0.34) and (1.13, 0.20) .. (1.81, 0.13)
            .. controls (2.13, 0.09) and (2.31, 0.08) .. (2.5, 0.08)
            -- (2.5, 0.08) -- (0, 0.08) -- cycle;
        \draw[black!50, dashed, line width=0.3pt] (0, 0.08) -- (2.5, 0.08);
        \draw[black!50, line width=1pt]
            (0, 0.67) .. controls (0.5, 0.34) and (1.13, 0.20) .. (1.81, 0.13)
            .. controls (2.13, 0.09) and (2.31, 0.08) .. (2.5, 0.08);
        \draw[black!40, line width=0.3pt] (3.35, 0) rectangle (4.45, 0.7);
        \fill[blue!25!black!30] (3.50, 0.05) rectangle (3.75, 0.55);
        \fill[blue!60!black!70] (4.05, 0.05) rectangle (4.30, 0.50);
        \node[font=\footnotesize\sffamily, text=black!60, anchor=north] at (3.625, 0.04) {ID};
        \node[font=\footnotesize\sffamily, text=black!60, anchor=north] at (4.175, 0.04) {OOD};
    \end{scope}

    \begin{scope}[yshift=0cm]
        \node[font=\footnotesize\sffamily, text=black!75] at (1.25, 0.95) {Noise};
        \draw[black!40, line width=0.3pt] (0, 0) rectangle (2.5, 0.7);
        \fill[orange!65!red, opacity=0.40]
            (0, 0.49) .. controls (0.625, 0.48) and (1.625, 0.45) .. (2.5, 0.44)
            -- (2.5, 0.44) -- (0, 0.44) -- cycle;
        \draw[black!50, dashed, line width=0.3pt] (0, 0.44) -- (2.5, 0.44);
        \draw[black!50, line width=1pt]
            (0, 0.49) .. controls (0.625, 0.48) and (1.625, 0.45) .. (2.5, 0.44);
        \draw[black!40, line width=0.3pt] (3.35, 0) rectangle (4.45, 0.7);
        \fill[blue!25!black!30] (3.50, 0.05) rectangle (3.75, 0.18);
        \fill[blue!60!black!70] (4.05, 0.05) rectangle (4.30, 0.15);
        \node[font=\footnotesize\sffamily, text=black!60, anchor=north] at (3.625, 0.04) {ID};
        \node[font=\footnotesize\sffamily, text=black!60, anchor=north] at (4.175, 0.04) {OOD};
    \end{scope}

    \node[font=\footnotesize\sffamily, text=black!60] at (1.25, -0.25) {Training tokens};
    \node[font=\footnotesize\sffamily, text=black!60, rotate=90] at (-0.4, 1.65) {Train loss};
    \node[font=\footnotesize\sffamily, text=black!60, rotate=90] at (2.92, 1.65) {Performance};
\end{scope}

\begin{scope}[xshift=6.05cm, yshift=1cm, local bounding box=panelB]
    \node[panel title] at (1.75, -1.7) {(b) Data selection};

    \node[dombox] (d1) at (0.5, 2.20) {Wikipedia};
    \node[dombox] (d2) at (0.5, 1.70) {ArXiv};
    \node[dombox] (d3) at (0.5, 1.20) {GitHub};
    \node[dombox] (d4) at (0.5, 0.70) {PubMed};

    \node[model] (learnerB) at (3.1, 1.45) {Learner\\[1pt]$P_\theta$};

    \draw[epi arrow, line width=1.7pt, opacity=0.9] (d1.east) -- ([yshift=4pt]learnerB.west);
    \draw[epi arrow, line width=0.7pt, opacity=0.55] (d2.east) -- ([yshift=1pt]learnerB.west);
    \draw[epi arrow, line width=1.3pt, opacity=0.8] (d3.east) -- ([yshift=-2pt]learnerB.west);
    \draw[epi arrow, line width=0.5pt, opacity=0.4] (d4.east) -- ([yshift=-5pt]learnerB.west);

    \node[slaw] (slaw) at (1.5, -0.6) {Scaling law\\[1pt]$\hat L_m(n_1,\ldots,n_K)$};
    \draw[fb arrow, solid] (learnerB.south) to[out=270, in=0] (slaw.east);

    \draw[decorate, decoration={brace, amplitude=5pt, mirror}, black!55, line width=0.5pt]
        (-0.30, 2.43) -- (-0.30, 0.47);
    \coordinate (bracetip) at (-0.52, 1.45);
    \draw[epi arrow, line width=0.8pt]
        (slaw.west) to[out=180, in=270, looseness=1.4] (bracetip);
    \node[font=\footnotesize\sffamily, text=orange!50!black] at (-0.85, 1.05) {$\pi_k$};
\end{scope}

\begin{scope}[xshift=10.95cm, yshift=1cm, local bounding box=panelC]
    \node[panel title] at (1.75, -1.7) {(c) Data generation};
    \node[gen, minimum width=1.5cm, minimum height=0.75cm] (gen) at (0.2, 2.2) {Generator\\[1pt]$P_{\theta^g}$};
    \node[model, minimum width=1.5cm, minimum height=0.75cm] (learnerC) at (0.2, -0.4) {Learner\\[1pt]$\theta_t \to \theta_{t+1}$};
    \node[buffer, minimum width=1.1cm, minimum height=1.8cm] (buf) at (2.4, 0.4) {};
    \foreach \y/\op in {1.15/0.3, 0.85/0.45, 0.55/0.6, 0.25/0.75, -0.05/0.9, -0.35/1.0} {
        \fill[black!55, opacity=\op, rounded corners=1pt]
            (2.1, \y-0.08) rectangle (2.7, \y+0.08);
    }
    \node[font=\footnotesize\sffamily, anchor=north] at (2.4, -0.6) {Buffer $\mathcal{B}$};
    \coordinate (xtbranch) at (gen.south |- buf.west);
    \draw[data arrow, -] (gen.south) -- node[right, font=\footnotesize\sffamily, text=black!60, pos=0.4] {$\mathcal{X}_t$}(xtbranch);
    \draw[data arrow] (xtbranch) --  (buf.west);
    \draw[data arrow] (xtbranch) -- (xtbranch |- learnerC.north);
    \fill[black] (xtbranch) circle (0.04);

    \draw[data arrow] (learnerC.east) -- (buf.west |- learnerC);
    \node[estbox, minimum width=1.30cm, minimum height=1.70cm] (estC) at (4.35, -0.10) {};
    \fill[blue!25!black!30] (3.95, -0.5) rectangle (4.2, 0.45);
    \node[font=\scriptsize\sffamily, text=black!70, anchor=north] at (4.055, -0.54) {$L_t^{\mathrm{pre}}$};
    \fill[blue!25!black!30] (4.35, -0.5) rectangle (4.6, 0.2);
    \node[font=\scriptsize\sffamily, text=black!70, anchor=north] at (4.675, -0.50) {$L_t^{\mathrm{post}}$};
    \draw[black!60, line width=0.3pt]
        (4.025, 0.45) -- (4.025, 0.6) -- (4.45, 0.6) -- (4.45, 0.2);
    \node[font=\footnotesize\sffamily, text=orange!50!black, anchor=west] at (4.5, 0.4) {$r_t$};

    \draw[data arrow, line width=0.9pt] (buf.east |- learnerC) -- (estC.west |- learnerC);

    \draw[epi arrow, line width=1.3pt]
        (estC.north) to[out=90, in=0, looseness=0.8]
        (gen.east);
\end{scope}

\begin{scope}
    \fill[orange!65!red, opacity=0.40] (6.7, 3.85) rectangle (7.15, 4.15);
    \draw[orange!65!red, opacity=0.40, line width=0.3pt] (6.7, 3.85) rectangle (7.15, 4.15);
    \node[anchor=west, font=\footnotesize\sffamily, text=black!70] at (7.25, 4.00) {Epiplexity signal};
\end{scope}

\end{tikzpicture}
}
\vspace{-0.3cm}
\caption{\textbf{Epiplexity for data selection and generation.} (a) Epiplexity, $S(t)$, can be estimated as the area between the learner's training loss curve and its final loss as a function of training tokens. Data which is trivial to learn or unlearnable will yield low estimates of epiplexity and result in lower performance in OOD downstream tasks. (b) For data selection, we use scaling laws to forecast the per-domain loss curves and predict the change in epiplexity, $\partial \hat{S}(t)/\partial n_j$, as a function of in-domain tokens. We then greedily maximize $\partial \hat{S}(t)/\partial n_j$ to determine the sampling weights $\pi_k$ used in adaptive reweighting of the domains during data selection. (c) For data generation, we estimate $d \hat{S}(t)/d t$ as the learner's loss reduction over an evaluation buffer $\mathcal{B}$, which stores the past training data. We set $d \hat{S}(t)/d t$ as the reward and update the generator via REINFORCE to steer the generator toward producing epiplexity-maximizing synthetic data.
}
\label{fig:ADAGE}
\end{figure}

\section{Background} \label{sec:background}
Our work integrates a diversity of research areas, including information theory and epiplexity, data selection,
scaling laws, self-play, and measures for tracking learning progress.

\textbf{Epiplexity} \quad
Epiplexity \citep{Finzi2026FromET} is a measure of the structural information in data that can be extracted by a model. Figure \ref{fig:ADAGE}(a) builds intuition in showing that this quantity is small when the data is trivial to learn (rapid convergence to low loss), as in the case of simple repetitive sequences, and when the data is unlearnable (no loss decrease), as in the case of random noise. In both conditions, the data contains little to no information to help make future predictions. Conversely, this quantity is large when the data contains substantial structural information which can be gradually internalized by a model (sustained loss decrease). While natural images or highly informative text may be relatively incompressible, they contain rich structural information that enables strong downstream predictions. 

The formal definition of epiplexity is the size of the model that minimizes the description length of data under a fixed compute budget. Computational constraints are important to the definition of epiplexity, because whether or not something is learnable depends on the computational resources (for example, while pseudorandom numbers are deterministically generated, they are indistinguishable from actual random numbers if we only have polynomial-time computation). More concretely, suppose ${P}_T$ is a probabilistic model that has a computational budget of $T$.
The time-bounded minimum description length (MDL) representation of data is given by the size of the model, $|P_T|$,  and the size of the data $X$ given the model, $\mathbb{E}[\log 1 / P_T\left(X\right)]$. Searching over the space of programs with a fixed computational budget $\mathcal{P}_T$, the model which minimizes this two part MDL is $P_{T}^{\star} =  \arg\min_{P_{T} \in \mathcal{P}_{T}} \left( \left|P_{T}\right| + \mathbb{E}[\log 1 / P_T\left(X\right)] \right)$. \emph{Epiplexity} is precisely the size of this program $S(T) = \left|P_{T}^{\star}\right|$.

\textbf{Estimating Epiplexity} \quad \citet{Finzi2026FromET} proposes using \emph{prequential coding} to estimate epiplexity via the training curve. Suppose a randomly initialized model $P_0$ is trained on a stream of i.i.d. sampled tokens $X_i$. The total entropy code length for the training data $X_{:M} = \{X_0, \dots, X_{M-1}\}$ and final model weights $P_M$ is given by $L(X_{:M},P_M) = \sum_{i=0}^{M-1} \log 1/P_i(X_i)$. Appealing to the symmetry of information, $L(Y) \approx L(X,Y) - L(X|Y)$, we can approximate the final model size $L(P_M)$, which is precisely our target definition of epiplexity, as the difference of  $L(X_{:M},P_M)$ and  $L(X_{:M}|P_M)$:
\vspace{-0.3em}
\begin{equation} \label{eq:prequential}
    |P_{\rm{preq}}| \approx \sum_{i=0}^{M-1} (\log1/P_i(X_i) - \log 1/P_M(X_i)) \,.
\end{equation}

The prequential estimate of epiplexity is given by Equation~\ref{eq:prequential}, which has a simple geometric interpretation. Each term can be seen as the gap between the loss incurred on a token at the moment it was seen and the loss incurred on that same token by the final model. These cumulatively sum to the area between the training loss curve and the floor set by the best model obtainable under the compute budget. We propose constructing tractable estimators for epiplexity atop this approximation of the measure. In Sections \ref{sec:data_selection} and \ref{sec:data_generation}, we introduce two novel estimation approaches, both of which replace the final loss with the current training loss, toward greedily maximizing epiplexity gain throughout training. 

\textbf{Data Selection via Scaling Laws} \quad
Data selection methods rest on the hypothesis that there exist curricula, or data orderings, that lead to improved model performance over random shuffles \citep{Xie2023DoReMiOD, Jiang2024AdaptiveDO, Liu2024RegMixDM, Chen2023SkillitAD, Gu2024DataSV, Lin2024Rho1NA, bengio2009Curriculum, mindermann2022prioritizedtrainingpointslearnable}. However, good curricula are hard to discover, relying on small model forecasting and expensive proxy training runs \citep{Xie2023DoReMiOD, Fan2023DoGEDR}, or meta-learning approaches which are computationally intractable \citep{Jiang2024AdaptiveDO, Gu2024TowardsOL}. Prior work has proposed using neural scaling laws to extrapolate model loss as a function of training data \citep{Jiang2024AdaptiveDO, Kaplan2020ScalingLF, Hoffmann2022TrainingCL}. For instance, a simple power law $\hat{L}(n;\alpha,\beta,\epsilon) = \epsilon + \beta n^{-\alpha}$ reasonably models loss behavior over training when fit online. \citet{Jiang2024AdaptiveDO} found that per-domain scaling laws can predict loss decreases and effectively guide data allocation across domains during training. However, this approach deferred to future work the modeling of cross-domain effects and instead relied on heuristic-based approximations. As the scaling laws trace the model's loss trajectories and epiplexity is measured as a function of that trajectory, this framework offers a natural opportunity to both apply epiplexity to the practical case of data selection among domains and to propose a parametric form for cross-domain scaling laws.

\begin{figure}
    \centering
    \begin{minipage}{0.32\linewidth}
        \includegraphics[width=\linewidth]{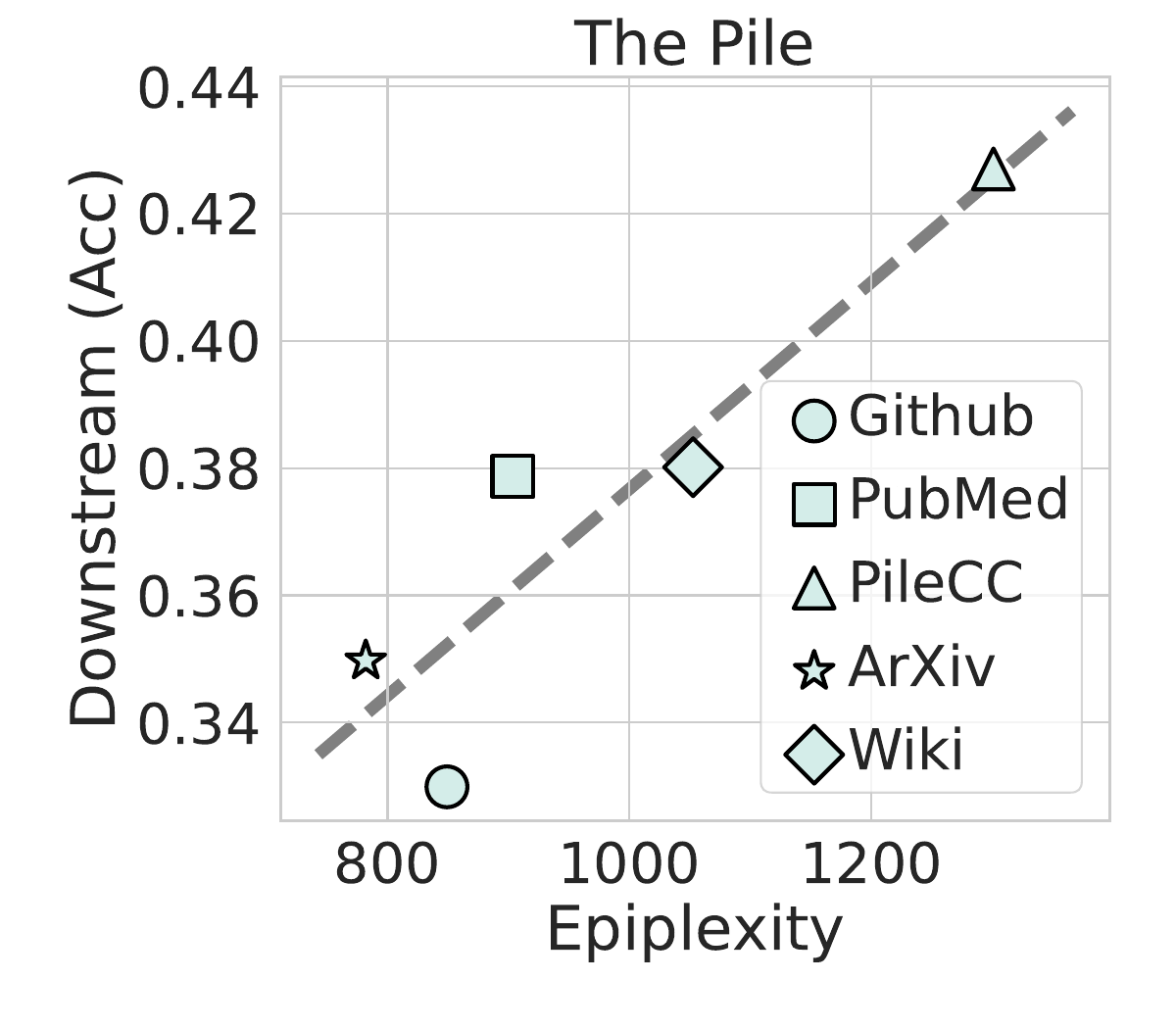}
    \end{minipage}
    \begin{minipage}{0.32\linewidth}
        \includegraphics[width=\linewidth]{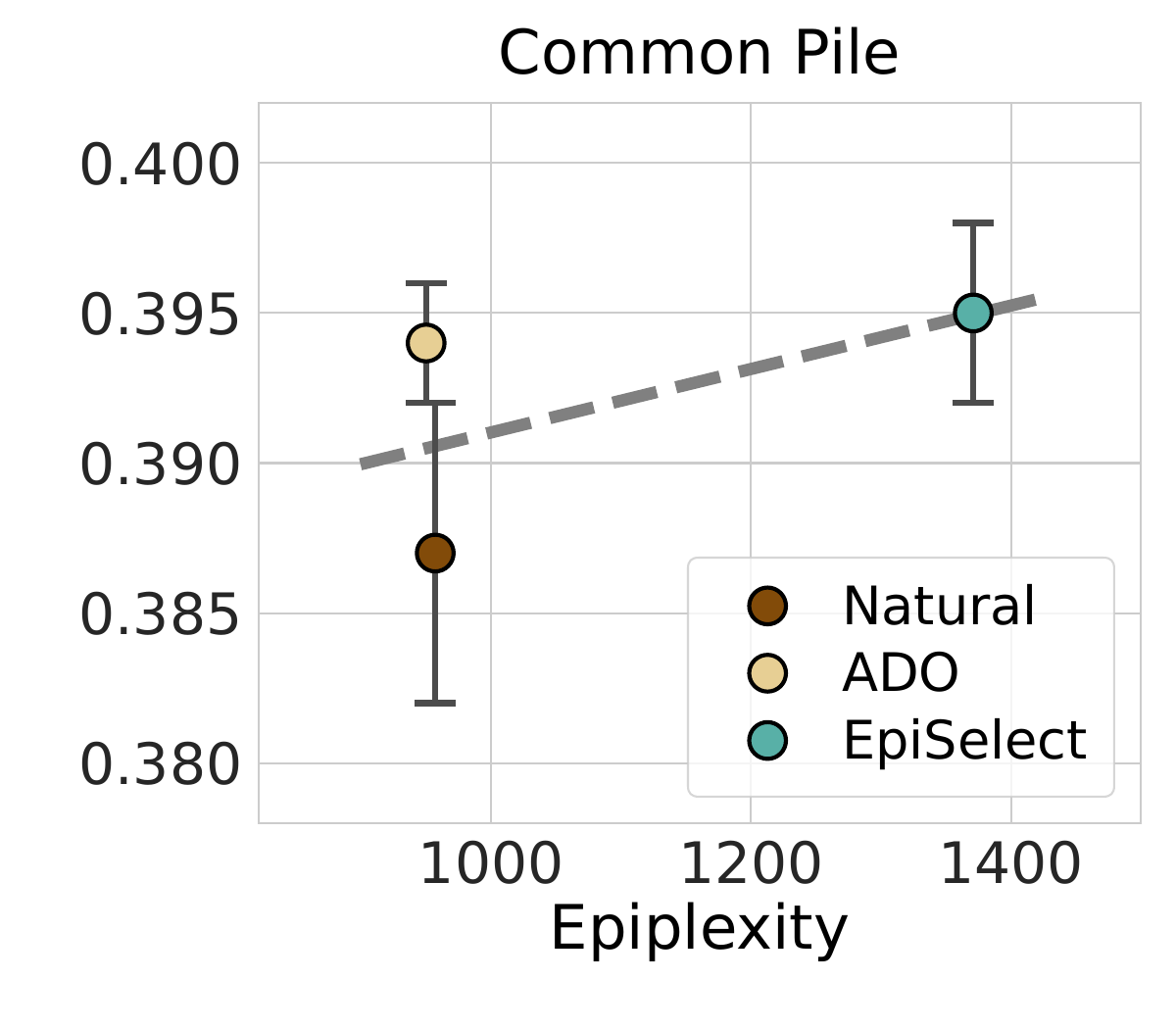}
    \end{minipage}
    \begin{minipage}{0.32\linewidth}
        \includegraphics[width=\linewidth]{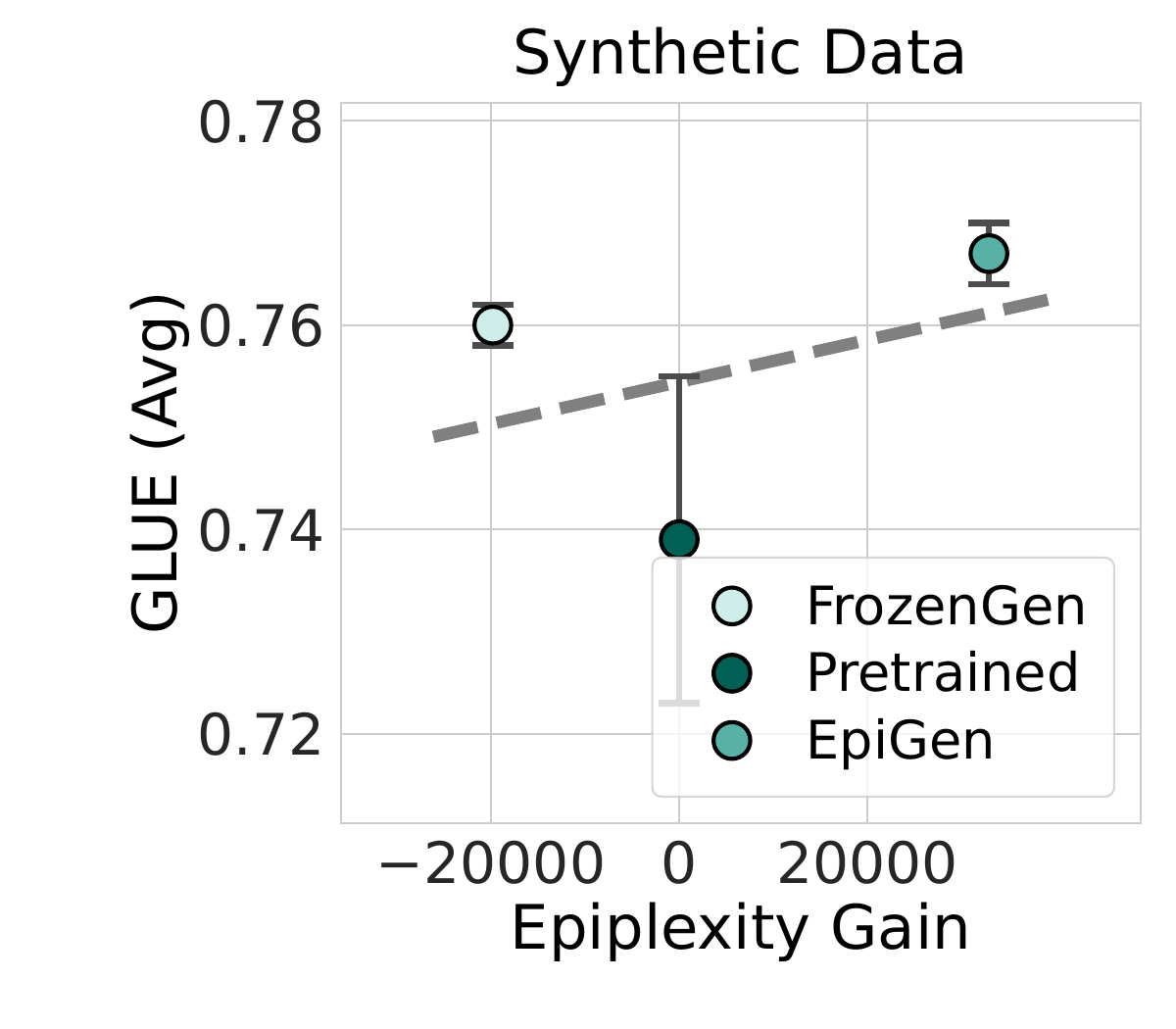}
    \end{minipage}
    \caption{
      \textbf{Higher epiplexity is correlated with better OOD generalization across data domains, data selection methods, and synthetic data generation.}
      In all plots, we contrast epiplexity against downstream performance; error bars show $\pm 1$ std over three seeds.
      \emph{(Left):} We estimate the epiplexity for the largest 5 domains on the Pile by solely training models on each domain and report average accuracy on a subset of the LM Eval Harness tasks.
      \emph{(Middle):} \texttt{EpiSelect} achieves higher epiplexity than alternatives and outperforms state-of-the-art curriculum learning procedures.
      We contrast the measured epiplexity and average zero-shot accuracy (Table~\ref{tab:downstream-std}, 124M) of \texttt{EpiSelect} on \texttt{Common Pile} against the \texttt{Natural} (no selection) baseline
      and \texttt{ADO} \citep{Jiang2024AdaptiveDO}, a SOTA method. 
      \emph{(Right):} Epiplexity guided synthetic data generation improves downstream performance.
      We plot the average GLUE score (Table~\ref{tab:glue_full}) of our initial checkpoint (\texttt{Pretrained}) and
      then compare the changes in epiplexity and performance when this checkpoint is trained on synthetic data naively generated from its original state (\texttt{FrozenGen})
      versus training on synthetic data from a generator trained to maximize epiplexity (\texttt{EpiGen}).
    }
    \label{fig:epiPOC}
\end{figure}

\textbf{Model Self-Play} \quad
Self-play has proven to support model improvement and learning without human supervision across many applications \citep{Tesauro1995, Silver2018AGR, Silver2016MasteringTG}. In self-play, two co-evolving agents each yield an automatic learning curriculum which tracks the opponent's current skill level, leading to mutual improvement. Recent works have applied self-play to language model alignment \citep{Chen2024SelfPlayFC, Yuan2024SelfRewardingLM} and capability improvement \citep{Liu2025SPIRALSO, Cheng2024SelfplayingAL, Zhao2025AbsoluteZR}. In particular, \citet{Liu2025SPIRALSO} demonstrated that self-play on zero-sum games can effect a 10\% improvement across reasoning benchmarks without the introduction of human-curated training data. The through-line goal of these works is for one agent to produce training data which is maximally informative---learnable, non-trivial, and that contains verifiable rewards---to the other agent at the current step in training. Epiplexity precisely targets this objective and offers a principled formalization of what self-play curricula aim to achieve.


\textbf{Tracking Learning Progress} \quad
Our implementation of epiplexity-guided synthetic data generation requires tracking the \textit{learning progress} of a model throughout training. Rewarding a generator by learning-progress metrics is a long standing approach, dating back to literature in artificial curiosity \citep{Schmidhuber2009DrivenBC, schmidhuber1991Curiosity, OudeyerCuriosity} and active learning \cite{Settles2009ActiveLL, houlsby2011bayesianactivelearningclassification}. These foundations have led to the broader paradigm of automatic curriculum learning, in which teacher models propose tasks at the limit of a learner model's competency \citep{Liu2025SPIRALSO,Zhao2025AbsoluteZR, Sukhbaatar2017IntrinsicMA, Matiisen2017TeacherStudentCL}. A failure mode in these approaches is that a generator rewarded purely for challenging the learner could collapse onto adversarial or low-entropy outputs that the learner memorizes without gaining generalizable structure \citep{Schmidhuber2009DrivenBC}. To mitigate this issue, prior work proposes verifiable environments (code executor, physical constraints in robotics, or surrogate difficulty tasks) to ground the rewards in external structure \citep{Liu2025SPIRALSO, Zhao2025AbsoluteZR,OudeyerCuriosity, Matiisen2017TeacherStudentCL}. In our approach of estimating the prequential signal of epiplexity over a generation history buffer, we effectively translate the information-theoretic quantity of epiplexity into a stable REINFORCE objective while obviating the need for external verifiers.

\section{EpiSelect: Maximizing Epiplexity to Select Data Domains} \label{sec:data_selection}

We introduce \texttt{EpiSelect}, a method of adaptively selecting batches of data which maximize the epiplexity through training. 

\subsection{Higher Epiplexity Predicts Better Downstream Generalization}
\label{sec: motivation}

We first motivate the design of our data selection method by demonstrating that higher epiplexity data domains correspond to improved downstream generalization. We train separate 124M LLaMA 2 family architecture \citep{touvron2023llama2openfoundation} models for 15B tokens on a few size-dominant domains (arXiv, GitHub, PileCC, PubMed, and Wikipedia) in The Pile \citep{Gao2020ThePA}, a popular benchmark for curriculum learning procedures. We limit this analysis to the five largest Pile domains since smaller domains lack sufficient in-domain tokens to train without multiple epochs (which violates the prequential estimator's assumption in Equation~\ref{eq:prequential} that credits code-length reduction to unseen tokens). We compute a post-hoc estimate of the epiplexity of each domain using the prequential estimator (Equation~\ref{eq:prequential}) and evaluate each model on a suite of 7 zero-shot tasks (\texttt{ARC}, \texttt{HellaSwag}, \texttt{LogiQA}, \texttt{WinoGrande}, \texttt{LAMBADA}, \texttt{SciQ} and \texttt{PIQA}) in the Language Model Evaluation Harness \citep{eval-harness}. As seen in Figure~\ref{fig:epiPOC} (\emph{Left}), epiplexity predicts average downstream performance across data domains (Pearson $r = 0.88$, two-tailed $p = 0.05$). In contrast, we found the model's weight norm, another scalar summary of the trained checkpoint, to show no such relationship ($r = 0.01$, Appendix~\ref{app:EpiCorrelation}). This suggests that epiplexity captures a meaningful signal about the transferability of training data, rather than a generic property of the trained model, and motivates using it directly as a selection objective.

\subsection{Tractably Estimating Per-Domain Epiplexity} \label{sec:epiSelect}
While standard training uses random minibatches of data, data selection methods pursue \emph{curriculum learning} to adaptively choose the domain distribution for each batch throughout training. Here, we propose a data selection procedure, \texttt{EpiSelect}, which computes a distribution $\pi \in \Delta^{K-1}$ over some $K$ data domains in a given dataset in order to maximize the expected epiplexity gain during training. Following the prequential estimator of epiplexity (Equation \ref{eq:prequential}), we estimate the total epiplexity $\hat{S}$ at training step $t$ with Equation \ref{eq:dataSelectTotalEpi}, where $n_{k}^{(s)}$ is the total number of domain $k$ tokens seen at iteration $s$, and $L_{m}(s)$ is the loss of domain $m$ at iteration $s$. 
\vspace{-0.3em}
\begin{equation}
    \hat{S}(t) = \sum_{m=1}^{K} \hat{S}_{m}(t) = \sum_{m=1}^{K} \sum_{s=1}^{t} (L_m(s) - L_m(t)) \label{eq:dataSelectTotalEpi} 
\end{equation}

In order to make online predictions for epiplexity gain during training, we draw inspiration from prior work \citep{Jiang2024AdaptiveDO} on single-domain scaling laws to propose a novel cross-domain scaling law in Equation \ref{eq:xDomainScalingLaw}, which predicts the loss on each domain $m$ as a function of the current token counts seen from every domain. During training, we repeatedly fit the scaling law for each domain $\hat{L}_{m}$ to the observed per-domain losses $L_m$ and per-domain tokens counts $n_m$ collected from the training trajectory so far, using Equation \ref{eq:xDomainScalingLaw}. Here, $\alpha_{m}$ and $\beta_m$ are positive constants, $\epsilon_{m}$ the irreducible error, and $\gamma_{m,k} > 0$ (with $\sum_{k=1}^{K}\gamma_{m,k} =1$) the normalized influence of all the domains on domain $m$ (fitting details in Appendix~\ref{app:xDomain}). While our model includes $\gamma_{m,k}$ terms to capture cross-domain interactions, the single-domain scaling law can easily be recovered by setting $\gamma_{m,k} = 0$ for $k \neq m$. Substituting the scaling laws into Equation \ref{eq:dataSelectTotalEpi} and taking the derivative with respect to the number of tokens from domain $j$ (full derivation for a single term in Appendix \ref{app:dataSelectEpi}), Equation \ref{eq:epiXDomainStepwise} arrives at a proxy for how much additional structure would be gained by training on additional tokens from domain $j$ at time $t$.
\vspace{-0.3em}
\begin{align}
    \hat{L}_m&(n_1, \dots, n_K) = \epsilon_m + \beta_m (\sum_{k=1}^{K}  \gamma_{m,k} n_k)^{-\alpha_m} \label{eq:xDomainScalingLaw} \\
    \frac{\partial \hat{S}}{\partial n_k} = \sum_{m=1}^{K} &\frac{\partial \hat{S}_m}{\partial n_k}
    = \sum_{m=1}^{K} \frac{n_m\alpha_{m} \gamma_{m,k}}{\sum_{i=1}^{K} \gamma_{m,i}{n_i}} (\hat{L}_m(n_1, \dots, n_K) - \epsilon_m) \label{eq:epiXDomainStepwise}
\end{align}

To maximize the overall epiplexity, we greedily maximize $\frac{\partial \hat{S}(t)}{\partial n_{k}}$ to select more tokens from domains which are predicted to contribute a greater increase in epiplexity. In Equation \ref{eq:piK}, we compute the distribution over domains, $\pi_k$, as a softmax over $\frac{\partial \hat{S}(t)}{\partial n_k}$, scaled by temperature parameter $\tau$, which allows for interpolation between the uniform distribution $\tau \rightarrow \infty$ and greedy maximization of marginal epiplexity $\tau \rightarrow 0$. In our experiments, we fixed $\tau =1$. $\pi_k$ serves as the domain sampling weights when selecting the next batch of training data. The full \texttt{EpiSelect} algorithm is given in Appendix \ref{app:epiDataSelectAlg}.
\vspace{-0.3em}
\begin{align} \label{eq:piK}
    \pi_{k} \propto \exp\left(\frac{1}{\tau} \frac{\partial \hat{S}(t)}{\partial n_{k}} \right)
\end{align}

\subsection{Saturation of Data Selection Methods on Pile} \label{sec:pileFinding}
\begin{wrapfigure}[14]{r}{0.4\textwidth}
  \centering
  \includegraphics[width=0.35\textwidth]{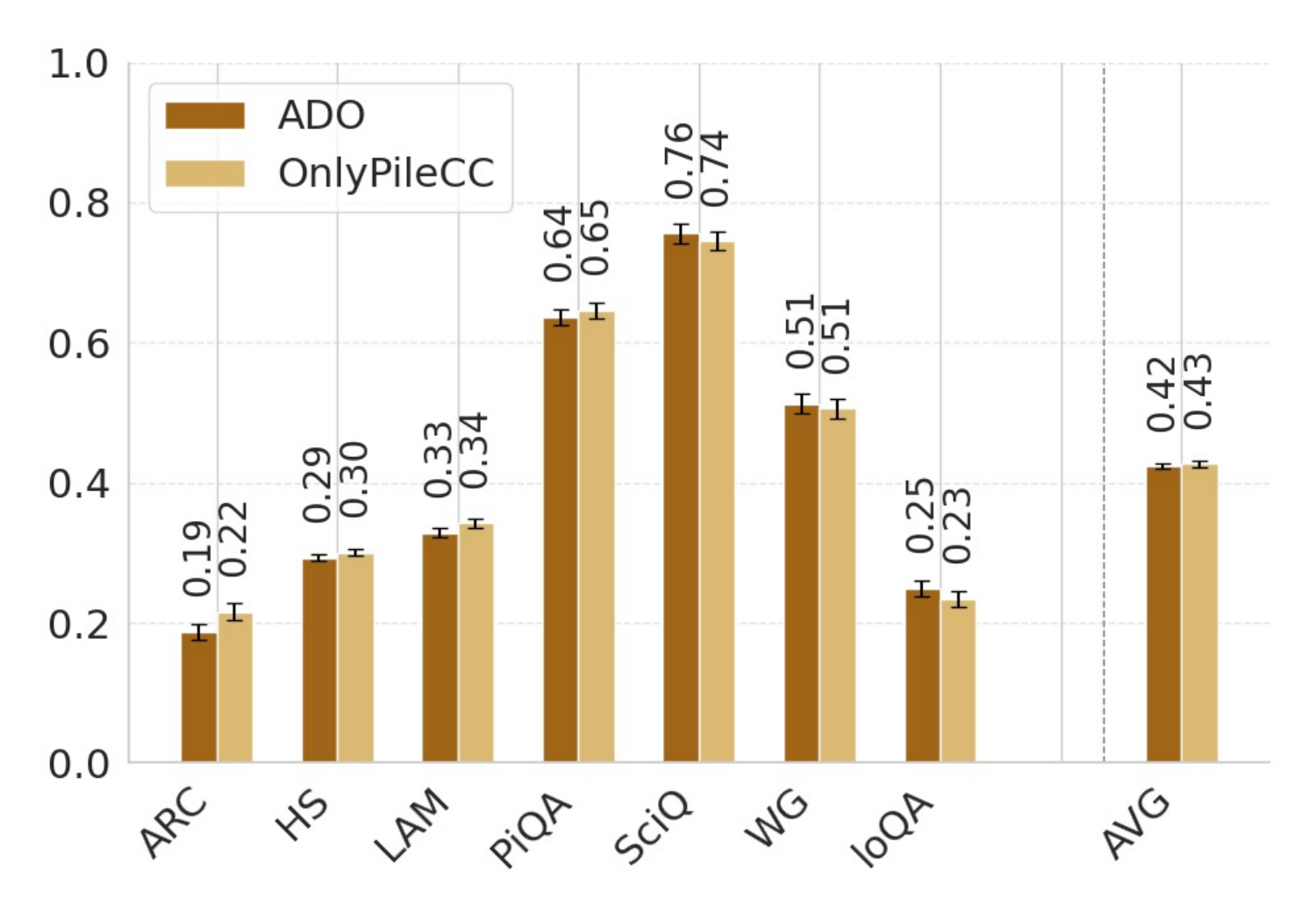}
    \caption{
    \textbf{The Pile does not require data selection.}
    Data selectors in the literature are trained to select domains on Pile, but solely training on the largest subdomain outperforms SOTA.
    }
  \label{fig:pileSaturation}
\end{wrapfigure}

We first trained 124M transformer models of the LLaMA 2 family \citep{touvron2023llama2openfoundation} on the uncopyrighted subset of the Pile dataset \citep{Gao2020ThePA}, which
contains around 210B tokens across 17 domains (excluding the original Books3, BookCorpus2, OpenSubtitles, YTSubtitles, and OWT2 domains due to copyright issues). However, due to the inherent domain size imbalance in the Pile uncopyrighted dataset, with one domain, PileCC, dominating at 41\% of the dataset tokens, we found that
the performance of data selection methods on Pile is saturated.
The performance across selection methods were highly similar and could be predicted by how each method selected from PileCC. We then trained our model exclusively on the PileCC domain, and Figure \ref{fig:pileSaturation} shows that this model was able to outperform \texttt{ADO} \citep{Jiang2024AdaptiveDO}, the SOTA selection method.
While it is valid that perhaps PileCC (a dataset of common crawl web text) is the most useful training data for model generalization, this result means that the
ideal selector would only ever select from the PileCC domain, which makes Pile uninformative as a benchmark for distinguishing between selection strategies.
Therefore, we chose to evaluate the selection methods on the newer Common Pile dataset, an 8TB collection of public-domain text.
Common Pile spans 30 domains, and the largest domain (Stack V2) accounts for roughly 14\% of the entire dataset. We built a token-balanced version of the Common Pile dataset (details in Appendix~\ref{app:CPTokenBalance}) and report subsequent data selection results on this construction.

\vspace{-2mm}
\subsection{Data Selection Experiments}
\label{sec: episelectresults}

We now show that by directly selecting for domains with high epiplexity, \texttt{EpiSelect} provides a practically compelling framework for curriculum learning. In these experiments, we train decoder-only transformer models (124M and 1.3B) of the LLaMA 2 family architecture \citep{touvron2023llama2openfoundation, Vaswani2017AttentionIA}
on the Common Pile dataset, an 8TB collection of public-domain text comprising 30 sources.
We compare our method (\texttt{EpiSelect}) against a no selection baseline (\texttt{Natural}), which samples batches proportionally to domain size (balanced according to the procedure in Appendix \ref{app:CPTokenBalance}), and the \texttt{ADO} selection method~\citep{Jiang2024AdaptiveDO}, a state-of-the-art approach for curriculum learning.
We train all models for a fixed budget of 15B tokens and evaluate all models on 10 common zero-shot downstream tasks in the Language Model Evaluation Harness \citep{eval-harness}. All results are reported as mean $\pm$ standard deviation over three independent runs, each with a different random seed.
Further details in Appendix~\ref{app:dataSelectImplementation}. Code available at \url{https://github.com/eysu35/EpiSelect}.

\newcommand{\pmstd}[1]{{\scriptsize$\pm$#1}}

\begin{table}[t]
\centering
\caption{
\textbf{\texttt{EpiSelect} matches or exceeds the strongest baseline at both model scales.}
Zero-shot accuracy (\%, higher is better) on LM Evaluation Harness tasks for 124M- and 1.3B-parameter models,
reported as mean $\pm$ standard deviation over 3 training seeds.
Highlighted cells indicate the top-performing method within each model family (all tied cells are highlighted).
We compare against checkpoints trained on the \texttt{Natural} data distribution and using the state-of-the-art
\texttt{ADO} selector \citep{Jiang2024AdaptiveDO}.
}
  \label{tab:downstream-std}
  \resizebox{\textwidth}{!}{
  \begin{tabular}{l *{11}{c}}
    \toprule
    &
      \textbf{ARC} & \textbf{BBQ} & \textbf{CSQA} & \textbf{HSwag} & \textbf{LAM} & \textbf{LogiQA2} & \textbf{OBQA} &
  \textbf{PIQA} & \textbf{SciQ} & \textbf{WinoG} & \textbf{Average} \\
    \midrule
    \multicolumn{12}{l}{\textit{124M}} \\
    \midrule
      \texttt{Natural} & 32.8\pmstd{0.5} & 40.6\pmstd{2.6} & 18.9\pmstd{0.9} & 28.0\pmstd{0.1} & 30.7\pmstd{0.7} &
        22.9\pmstd{1.3} & 15.7\pmstd{0.4} & 60.8\pmstd{0.4} & 85.5\pmstd{0.8} & 50.7\pmstd{1.7} & \textbf{38.7}\pmstd{0.5} \\
      \texttt{ADO} & 34.0\pmstd{0.3} & 40.1\pmstd{2.1} & \cellcolor{first} 20.5\pmstd{0.6} & 28.1\pmstd{0.1} &
        \cellcolor{first} 30.9\pmstd{0.7} & \cellcolor{first} 23.9\pmstd{0.4} & \cellcolor{first} 16.1\pmstd{0.5} & 62.3\pmstd{0.4} &
        \cellcolor{first} 86.5\pmstd{0.7} & \cellcolor{first} 51.7\pmstd{1.0} & \textbf{39.4}\pmstd{0.2} \\
      \texttt{EpiSelect} & \cellcolor{first} 34.9\pmstd{0.6} & \cellcolor{first} 43.0\pmstd{2.0} & 20.3\pmstd{0.9} &
        \cellcolor{first} 28.3\pmstd{0.1} & 30.2\pmstd{0.6} & 23.6\pmstd{0.3} & 15.5\pmstd{0.8} & \cellcolor{first} 62.4\pmstd{0.7} &
        \cellcolor{first} 86.5\pmstd{0.9} & 50.9\pmstd{0.9} & \cellcolor{first} \textbf{39.5}\pmstd{0.3} \\
    \addlinespace
    \midrule
    \multicolumn{12}{l}{\textit{1.3B}} \\
    \midrule
      \texttt{Natural} & 40.0\pmstd{0.6} & 32.3\pmstd{0.7} & 19.6\pmstd{0.7} & 33.2\pmstd{0.1} & 45.2\pmstd{0.5} &
        24.1\pmstd{0.5} & 19.1\pmstd{0.2} & 65.3\pmstd{0.3} & 91.1\pmstd{0.7} & 52.2\pmstd{0.6} & \textbf{42.2}\pmstd{0.2} \\
      \texttt{ADO} & 41.1\pmstd{0.5} & 31.5\pmstd{1.4} & 20.1\pmstd{1.5} & \cellcolor{first} 34.3\pmstd{0.2} & \cellcolor{first} 45.5\pmstd{0.5} &
        24.6\pmstd{1.0} & \cellcolor{first} 20.3\pmstd{1.5} & 67.3\pmstd{0.9} & 91.2\pmstd{0.5} & 52.3\pmstd{0.6} & \textbf{42.8}\pmstd{0.2} \\
      \texttt{EpiSelect} & \cellcolor{first} 42.0\pmstd{0.4} & \cellcolor{first} 33.4\pmstd{0.2} & \cellcolor{first} 20.3\pmstd{0.5} &
        \cellcolor{first} 34.3\pmstd{0.2} & \cellcolor{first} 45.5\pmstd{0.7} & \cellcolor{first} 24.9\pmstd{1.2} & 19.7\pmstd{0.2} & \cellcolor{first}
  67.8\pmstd{0.5} &
        \cellcolor{first} 91.6\pmstd{0.3} & \cellcolor{first} 53.2\pmstd{1.1} & \cellcolor{first} \textbf{43.3}\pmstd{0.1} \\
    \bottomrule
  \end{tabular}
  }
  \end{table}

\textbf{Epiplexity Maximizing Selector Achieves Highest Performance Across Zero-Shot Tasks} \quad
We evaluate our epiplexity maximizing selector on a set of 10 zero-shot language understanding tasks from the LM Evaluation Harness \citep{eval-harness}: ARC (ARC-Easy and ARC-Challenge, averaged), BBQ, CSQA, HellaSwag (HSwag), LAMBADA (LAM), LogiQA2, OBQA, PIQA, SciQ, and WinoGrande (WinoG). In Table \ref{tab:downstream-std}, we show that for both model sizes, 
\texttt{EpiSelect} achieves the highest average task accuracy (39.5\% and 43.3\% for 124M and 1.3B, respectively),
matching \texttt{ADO} at 124M and outperforming both the \texttt{Natural} 
(no selection) baseline and \texttt{ADO} at 1.3B by margins comparable to those reported in prior work~\citep{Jiang2024AdaptiveDO}.
The improvements grow with scale, with \texttt{EpiSelect} attaining or tying for the top score on 5 of 10 tasks at 124M and 9 of 10 tasks at 1.3B. These results provide the initial support for our central hypothesis: by reweighting the batches toward domains predicted to contribute the largest marginal increase in epiplexity,
the selector induces a training curriculum that produces representations which transfer more effectively to a diverse set of OOD tasks.
\begin{wrapfigure}[13]{l}{0.25\textwidth}
  \centering
  \includegraphics[width=0.25\textwidth]{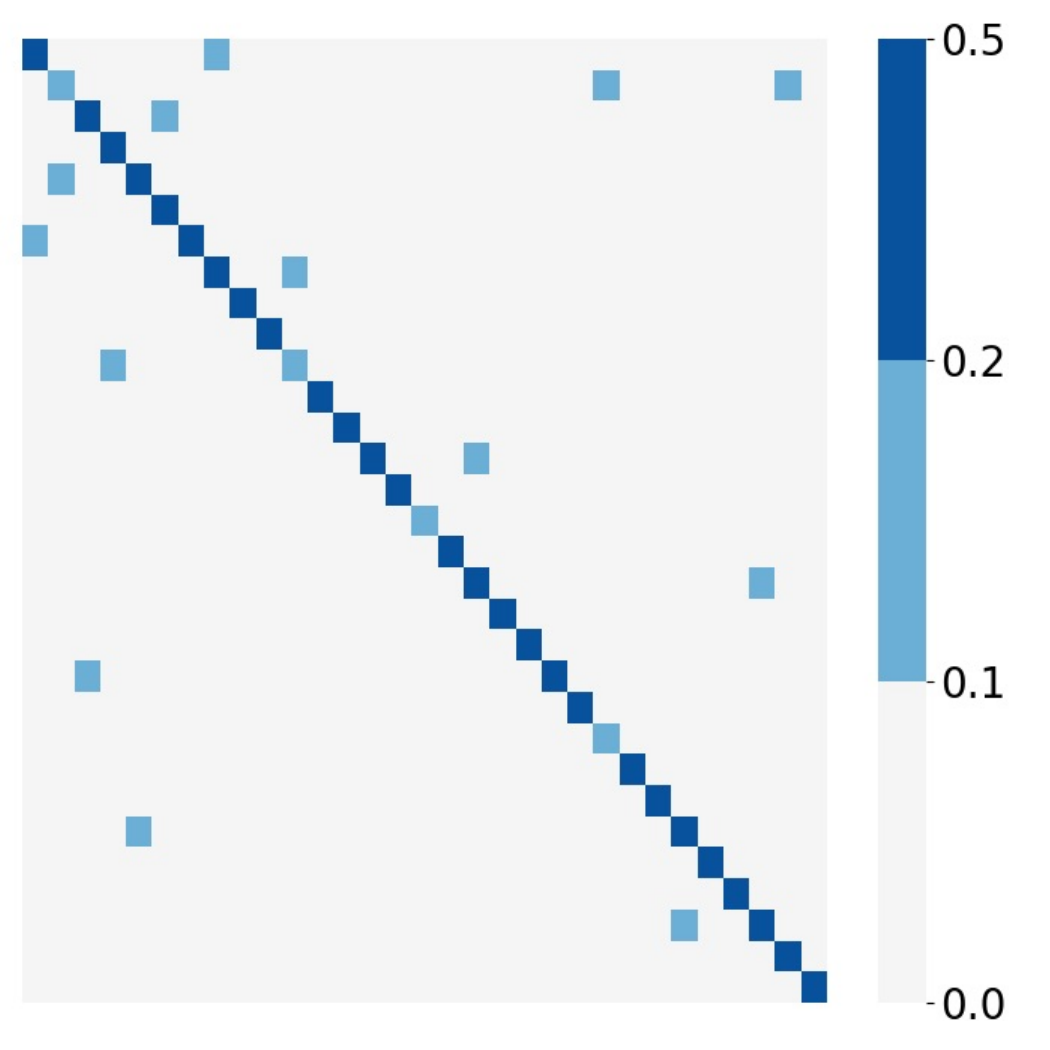}
  \caption{
    \textbf{Few domains interact.}
    Diagonal $\gamma_{m,k}$ values dominate.
  }
  \label{fig:gammas}
\end{wrapfigure}

\textbf{Cross-Domain Effects Are Sparse and Directional} \quad

A key component of estimating epiplexity correctly is to model the cross-domain effects as seen in Equation \ref{eq:xDomainScalingLaw}.
Figure \ref{fig:gammas} plots the matrix of fitted $\gamma_{m,k}$ values at the end of training, which captures the weighting that domain $k$ tokens have when computing
the predicted loss in domain $m$. We see that, beyond improved downstream performance, our cross-domain scaling laws surface an interpretable map of inter-domain influence.
By the end of training, the fitted $\gamma_{m,k}$ matrix is diagonally dominant, which aligns with our intuition that training on domain $k$ tokens
should have the highest impact toward reducing loss on domain $k$.
We also observe a sparse set of strong cross-domain effects, in the off-diagonal values, and note an asymmetry in the domain interactions.

\section{EpiGen: Maximizing Epiplexity for Synthetic Data Generation} \label{sec:data_generation}

Next, considering cases when a training dataset is not neatly divided into domains or we require additional training data, we turn to synthetic data \textit{generation}. For the first time, we show how epiplexity can be operationalized to select over a generator model's weights to effectively guide synthetic data generation. Intuitively, synthetic training data should challenge the learner, leading to sustained and substantial decreases in loss, while not being trivial or noise. As previewed in panel (a) of Figure \ref{fig:ADAGE}, the prequential coding estimate of epiplexity captures this notion exactly and is well-suited to guide a generator model toward producing optimal synthetic data.

We train a generator model, $P_{\theta^g}$, to generate high-epiplexity synthetic data used to train a separate learner model, $P_{\theta}$. To estimate the epiplexity for the learner model, we implement a generation buffer $\mathcal{B}$, which accumulates the generated data throughout training, including the current batch $\mathcal{X}_t$.
Our use of a buffer differs in function from experience replay \citep{Mnih2015HumanlevelCT, Lin1992SelfimprovingRA}, where a buffer stores past transitions for off-policy training updates.
In our method, the buffer is not used to train either model. Instead, it serves as a reference distribution for reward computation to provide a tractable estimate of epiplexity.
For each batch of generated data, Equation \ref{eq:epiDataGen} estimates the change in epiplexity as the difference between the learner loss on the samples in the buffer before ($P_{\theta_{t-1}}$) and after ($P_{\theta_{t}}$) training on the current batch.
\vspace{-0.3em}
\begin{align}
r_t = S(t) &- S(t-1) = \sum_{x \in \mathcal{B}} \left( \log(1/P_{\theta_{t-1}}(x)) - \log(1/P_{\theta_{t}}(x)) \right) \label{eq:epiDataGen}\\
\nabla_{\theta^{g}} &\mathcal{J} = \mathbb{E}_{\mathcal{X}_t \sim P_{\theta^{g}}} [(r_t - b) \cdot \sum_{x \in \mathcal{X}_t}\nabla_{\theta^{g}} \log
P_{\theta^{g}} (x)] \label{eq:reinforceObj}
\end{align}

In order for the generator to learn to maximize the resulting epiplexity, we set the reward $r_t$ for the model as this step-wise change in epiplexity and optimize the objective $\mathcal{J}(\theta^{g}) = \mathbb{E}[r_t]$ using REINFORCE policy gradients \citep{Sutton1999PolicyGM}. In Equation \ref{eq:reinforceObj}, the gradient can be expressed as a product of an advantage term (baseline $b$ tracks the reward history with an exponential moving average) and the log-likelihood of the batch. 

\subsection{Synthetic Data Generation Experiments}\label{sec:datagenResult}
To empirically validate \texttt{EpiGen}, we initialize a generator and a learner model from pretrained GPT-2 weights \citep{radford2019gpt2}. 
We iteratively generate batches of synthetic data from the current generator model, train the learner for some $K$ steps on this data, then train the generator for one step via REINFORCE policy gradients to optimize the generator for an epiplexity-maximizing distribution. The full \texttt{EpiGen} algorithm and implementation details are respectively provided in Appendices \ref{app:epiDataGenAlg} and \ref{app:dataGenImplementation}. Code available at \url{https://github.com/eysu35/EpiGen}. 

In developing our method, we found that the epiplexity-based reward leads to REINFORCE gradients with low variance and overall stable generation dynamics (shown in Appendix \ref{app:datagenTrainingDynamics}). We thus compared the trained \texttt{EpiGen} learner model against the pretrained GPT-2 checkpoint (\verb|Pretrained|) and a set of baseline learner models trained by the following generators: a frozen generator to match the total token count (\verb|FrozenGen|),
a generator trained with a heuristic-based reward (\verb|PPL|, negative log perplexity), and a generator trained with a learning-based reward that only considers the current batch of data (\verb|NoBuffer|). All results are reported as mean $\pm$ standard deviation over three independent runs of the full training pipeline, each with a different random seed.

\textbf{Epiplexity Reward Yields Learnable and Natural Synthetic Data} \quad
Our first evaluation of \texttt{EpiGen} examines learner shift from the pretrained distribution and the quality of the generated synthetic data. For the
former, we evaluate the initial learner (pretrained checkpoint) against all learners trained on synthetic data from the various generators on in- and
out-of-distribution natural language datasets: OpenWebText \citep{Gokaslan2019OpenWeb} (OWT), an open source reproduction of GPT-2's pretraining dataset
\citep{radford2019gpt2}, SlimPajama \citep{soboleva2023slimpajama}, and FineWeb \citep{penedo2024fineweb}. For the latter, we sample 256 512-token sequences from each final generator and measure their perplexity under pretrained GPT-2, the generator's sampling entropy, and their within-sequence 4-gram repetition rate and compare these metrics against those of 256 real OpenWebText samples. Additional evaluation details are in Appendix~\ref{app:oodBenchmarks} and Appendix~\ref{app:generatorStats}.

From Table \ref{tab:epigenTAB1}, we draw three conclusions. First, any training on synthetic data increases perplexity on natural language. Since synthetic
batches are randomly sampled from the generator at each step, this data is unsurprisingly ungrammatical, so even the \texttt{FrozenGen} learner drifts from
its pretrained starting point (24.7 to 26.6 on OWT, 29.8 to 31.2 on SlimPajama, and 35.1 to 37.7 on FineWeb). Second, among the rewards that train the
generator, the \texttt{EpiGen} learner best predicts natural language (OWT PPL of 27.4 vs.\ 32.9 and 29.0), and the same ordering holds out of distribution
(SlimPajama 32.6 vs.\ 38.4 and 34.4; FineWeb 39.4 vs.\ 48.9 and 41.8), suggesting that the epiplexity-based signal guides the generator toward samples
similar to real text. Third, statistics of the generated samples (GPT-2 perplexity, entropy in nats, and 4-gram repetition rate) show that the
epiplexity-based reward is the only one which maintains natural-language similarity. The \texttt{PPL} generator achieves the lowest GPT-2 ppl of $5.8$, but
its samples are highly repetitive (4-gram rep.\ of $0.468$ vs.\ \texttt{EpiGen}'s $0.007$) and thus more trivial to learn. Conversely, the \texttt{NoBuffer}
generator, trained directly on learner loss, produces samples with low repetition ($0.002$) but a high GPT-2 perplexity of $169.7$ (vs.\ \texttt{EpiGen}'s
$58.2$). Here, the generator has become adversarial. To maximize its reward, it produces samples that are highly unlikely under the learner, i.e., samples
that approach noise. In accordance with our hypothesis, formulating the generator's reward as an approximation of epiplexity is the only method which
effectively trains the generator while keeping its samples learnable and close to natural text. For further validation, we randomly sampled from the
synthetic batches during generation time and include \texttt{EpiGen} examples from steps 1000 (first sample) and 60,000 (last sample) in Appendix
\ref{app:syntheticDataEx}.

\begin{table}[t!]
    \centering
    \caption{
          \textbf{Epiplexity-based reward yields highest quality synthetic data.} \emph{Left:} perplexity of
          the pretrained GPT-2 checkpoint and of the learners trained on synthetic data from a frozen generator
          (\texttt{FrozenGen}), the epiplexity reward (\texttt{EpiGen}), and the alternative rewards
          (\texttt{PPL}, \texttt{NoBuffer}) on OpenWebText and two OOD corpora (lower is better). \emph{Right:}
          statistics of samples drawn from each final generator, with real OpenWebText text for reference.
          All values are mean $\pm$ std over three seeds, where seed $s$ sets both training and sampling.
          \texttt{FrozenGen}'s generator is pretrained GPT-2 in every seed, so its std is sampling variance
          alone. Highlighted cells mark the trained-generator method closest to the pretrained or real-text baseline.}
    \label{tab:epigenTAB1}
    \resizebox{\textwidth}{!}{
\begin{tabular}{l ccc ccc}
        \toprule
         & \multicolumn{3}{c}{\textbf{Learner perplexity} $\downarrow$} & \multicolumn{3}{c}{\textbf{Generator samples}} \\
        \cmidrule(lr){2-4} \cmidrule(lr){5-7}
         & \textbf{OWT} & \textbf{SlimPajama} & \textbf{FineWeb} & \textbf{GPT-2 ppl} & \textbf{Entropy (nats)} & \textbf{4-gram rep.} \\
        \midrule
        \textcolor{gray}{Real text} & \textcolor{gray}{--} & \textcolor{gray}{--} & \textcolor{gray}{--} &
\textcolor{gray}{25.4} & \textcolor{gray}{--} & \textcolor{gray}{0.034} \\
        \texttt{Pretrained} (no training)     & 24.7 & 29.8 & 35.1 & \textcolor{gray}{--} & \textcolor{gray}{--} & \textcolor{gray}{--}\\
        \texttt{FrozenGen} (generator frozen) & 26.6\pmstd{0.1} & 31.2\pmstd{0.1} & 37.7\pmstd{0.1} & 63.1\pmstd{0.0} & 4.16\pmstd{0.00} & 0.016\pmstd{0.000} \\
        \midrule
        \multicolumn{7}{l}{\textit{Generator trained}} \\
        \texttt{EpiGen}   & \cellcolor{first} 27.4\pmstd{0.1} & \cellcolor{first} 32.6\pmstd{0.1} & \cellcolor{first} 39.4\pmstd{0.1} & \cellcolor{first} 58.2\pmstd{2.1} & \cellcolor{first} 4.05\pmstd{0.04} &
\cellcolor{first} 0.007\pmstd{0.001} \\
        \texttt{PPL}      & 32.9\pmstd{0.4} & 38.4\pmstd{1.0} & 48.9\pmstd{1.2} & 5.8\pmstd{0.3} & 1.65\pmstd{0.06} & 0.468\pmstd{0.013} \\
        \texttt{NoBuffer} & 29.0\pmstd{0.6} & 34.4\pmstd{1.2} & 41.8\pmstd{1.2} & 169.7\pmstd{14.5} & 5.07\pmstd{0.11} & 0.002\pmstd{0.001} \\
        \bottomrule
    \end{tabular}
    }
\end{table}
\textbf{Training on Synthetic Data Improves Downstream Generalization} \quad
Next, we evaluate whether the synthetic data produced by our method improves the generalization of the learner. We follow the standard fine-tuning procedure on seven natural language understanding tasks from the GLUE benchmark \citep{Wang2018GLUEAM} (CoLA, SST-2, MRPC, QQP, MNLI, QNLI, RTE), updating all weights along with a linear classification head, and evaluate on the development sets (details in Appendix~\ref{app:dataGenImplementation}). We report each task's standard metric (Matthews correlation for CoLA, accuracy otherwise) and the average score. As shown in Table~\ref{tab:glue_full}, the \texttt{EpiGen} learner outperforms the pretrained baseline, improving the average score by 2.8 points, with gains driven by CoLA and RTE. Critically, none of the learners in Table~\ref{tab:glue_full} have seen more tokens of real data than the pretrained model, so these improvements come entirely from training on synthetic data. We note that this gain is not specific to \texttt{EpiGen}. Every learner trained on additional synthetic data improved over the pretrained checkpoint to some degree. While the differences between variants are small compared to the seed variance, we read the GLUE results as preliminary evidence that training on data sampled from the pretrained generator can improve downstream generalization. 

\begin{table}[t!]
    \centering
    \caption{
        \textbf{GLUE transfer for all learners.} Validation scores ($\times$100), mean $\pm$ std over three
        seeds. CoLA is Matthews correlation, the rest accuracy. Highlighted cells indicate the top-performing learner on each task. Every learner trained on synthetic data improves over the pretrained checkpoint by 2.0–2.8 average points, so the choice of generator should consider the learner and generator quality in Table~\ref{tab:epigenTAB1}.
    }
    \label{tab:glue_full}
    \resizebox{\textwidth}{!}{
    \begin{tabular}{l ccccccc c}
        \toprule
         & \textbf{CoLA} & \textbf{SST-2} & \textbf{MRPC} & \textbf{QQP} & \textbf{MNLI} & \textbf{QNLI} & \textbf{RTE} & \textbf{Average} \\
        \midrule
        \texttt{Pretrained} (no training)     & 26.1\pmstd{9.5} & 91.4\pmstd{0.2} & 77.0\pmstd{0.9} & 89.0\pmstd{0.2} & 81.6\pmstd{0.2} & 88.2\pmstd{0.1} & 64.3\pmstd{0.4} & \textbf{73.9}\pmstd{1.6} \\
        \texttt{FrozenGen} (generator frozen) & 38.6\pmstd{2.2} & 91.3\pmstd{0.4} & \cellcolor{first} 78.7\pmstd{1.1} & \cellcolor{first} 89.2\pmstd{0.1} & 81.6\pmstd{0.1} & 88.3\pmstd{0.3} & 64.5\pmstd{0.7} & \textbf{76.0}\pmstd{0.2} \\
        \midrule
        \multicolumn{9}{l}{\textit{Generator trained}} \\
        \texttt{EpiGen}   & \cellcolor{first} 41.5\pmstd{3.2} & \cellcolor{first} 91.6\pmstd{0.4} & 78.3\pmstd{0.4} & 89.1\pmstd{0.1} & 81.5\pmstd{0.2} & \cellcolor{first} 88.4\pmstd{0.2} & \cellcolor{first} 66.4\pmstd{0.7} & \cellcolor{first} \textbf{76.7}\pmstd{0.3} \\
        \texttt{PPL}      & 37.5\pmstd{2.6} & 91.1\pmstd{0.4} & 78.5\pmstd{1.0} & 89.1\pmstd{0.2} & \cellcolor{first} 81.7\pmstd{0.2} & 88.3\pmstd{0.3} & 65.0\pmstd{1.3} & \textbf{75.9}\pmstd{0.3} \\
        \texttt{NoBuffer} & 39.1\pmstd{2.6} & 91.4\pmstd{0.4} & 78.2\pmstd{0.0} & 89.1\pmstd{0.1} & 81.5\pmstd{0.2} & \cellcolor{first} 88.4\pmstd{0.2} & 66.3\pmstd{0.6} & \textbf{76.3}\pmstd{0.4} \\
        \bottomrule
    \end{tabular}
    }
\end{table}

\textbf{Mixtures of Synthetic and Real Data} \label{sec:additionalTraining}\quad
Although self-play paradigms are inherently useful in not requiring additional data to improve model performance, we conduct a data mixing experiment to assess how the epiplexity-guided synthetic data stack up against novel real data. 
\begin{wrapfigure}[20]{l}{0.48\textwidth}
  \centering
  \includegraphics[width=0.46\textwidth]{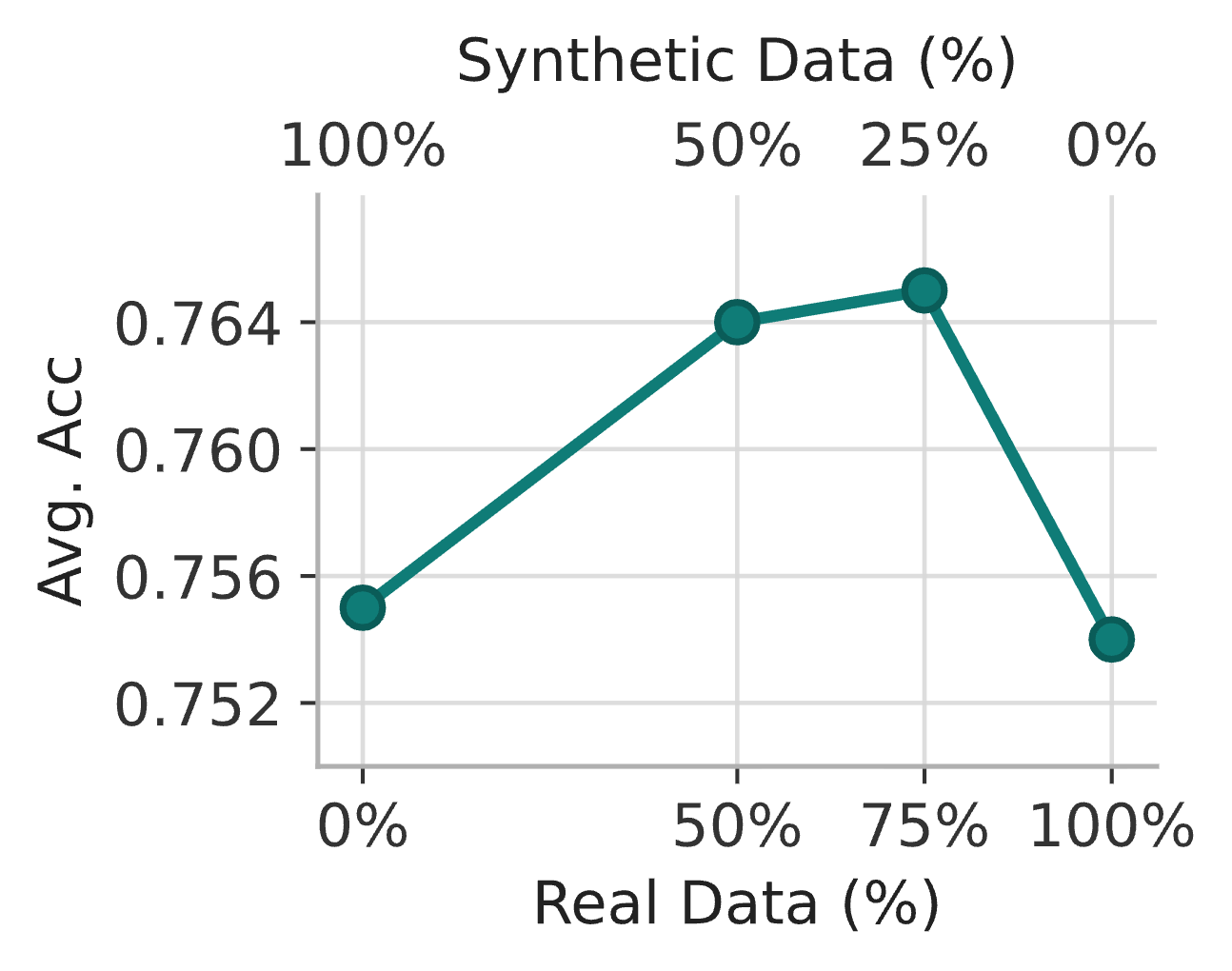}
  \caption{\textbf{Mixing synthetic and real data yields the greatest improvement}. Average GLUE score for learners trained on varying proportions of epiplexity-guided synthetic data and OWT data.
  }
  \label{fig:dataMixing}
\end{wrapfigure}
Here, we initialize the generator model as the \textit{trained}, epiplexity-maximizing generator from the experiments above. Then, we run the data generation pipeline with another learner model (initialized with the pretrained GPT-2 weights) and keep the generator weights frozen. In this way, we train the learner on the epiplexity-maximizing synthetic dataset for the entire duration of training, and can incrementally replace fractions of the batches of synthetic dataset with samples from the OWT dataset, all of the same sequence length, to further ablate the impact of synthetic data training on downstream performance. We make note of an additional analysis in Appendix \ref{app:OWTBuffer}, in which we ablate the fraction of OWT data in the evaluation buffer to more realistically approximate epiplexity over all prior training data, real and synthetic. Figure \ref{fig:dataMixing} shows that learner models trained on 50\% and 25\% of synthetic data (and, subsequently, 50\% and 75\% of OWT data) achieve higher average performance on the OOD GLUE benchmark, indicating that training on synthetic and real data outperforms training on either one individually. This result both further validates the performance improvement we observed in Table \ref{tab:glue_full} and substantiates the need for principled synthetic data generation methods.  

\begin{wrapfigure}[17]{r}{0.45\textwidth}
  \centering
  \includegraphics[width=0.4\textwidth]{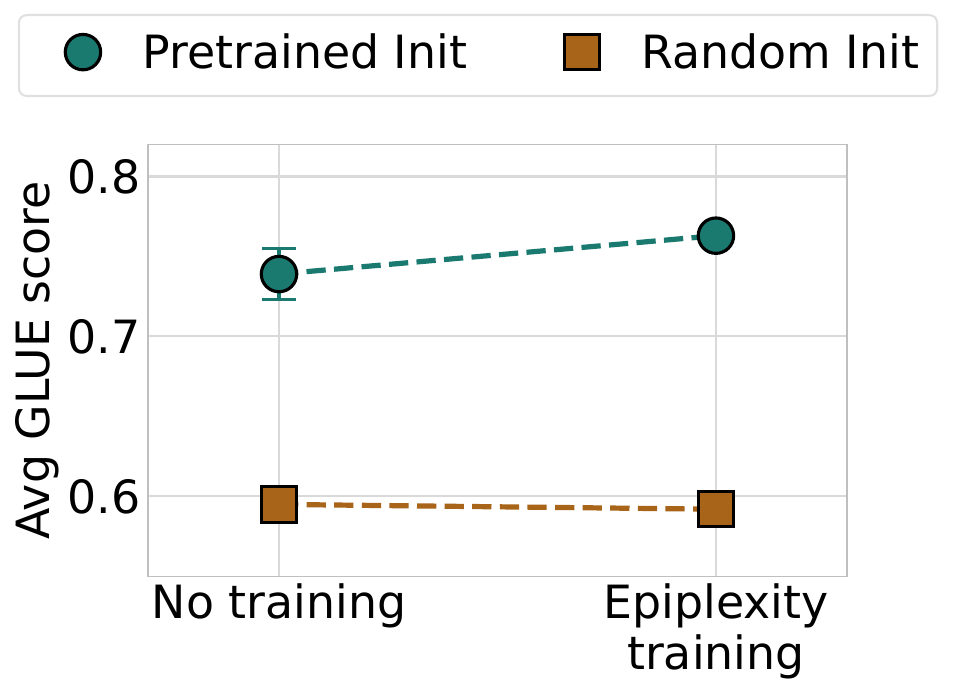}
    \caption{\textbf{Synthetic data learning signal emerges from pretrained model weights.} Average GLUE score for pretrained vs. random initialization, with and without epiplexity-guided training. Error bars show $\pm 1$ std over three seeds for the pretrained initialization (some smaller than the markers).
    }
  \label{fig:RandomInit}
\end{wrapfigure}

\textbf{The Importance of Pretrained Initialization} \quad
To better understand the efficacy of our method, we 
investigate the effect of model initialization to determine where the additional learning signal sources from. We ran our data generation pipeline with generator and learner model initialized with random weights and with pretrained GPT-2 weights.
Figure \ref{fig:RandomInit} shows that, while epiplexity-guided training improves average raw GLUE performance from 0.739 to 0.767 with the pretrained GPT-2 weights, it provides essentially no change (-0.003) with the random initialization. This finding is consistent with the observation that while the ceiling of reinforcement learning is in principle not bound to the pre-trained model, empirically the quality of the pretrained initialization still constrains the performance after RL \citep{yue2504does}. Finally, we showed in Appendix~\ref{app:OWTBuffer} that at matched token budgets, learners trained on epiplexity-guided synthetic data perform comparably to learners fine-tuned directly on the same number of real OpenWebText tokens (Appendix Figure~\ref{fig:OWTAblation}).

\section{Discussion}
Machine learning research has historically operated under the assumption that we wish to maximize generalization performance on
test points drawn from the same distribution as a set of training data. From the model selection side, this setting has led to principles such
as Occam's razor, where we wish to have the simplest explanation that is consistent with our observations. However, we are now entering uncharted territory, where we train models on heterogeneous data and deploy them in new unanticipated domains. This
setting forces a shift in perspective from model selection to data selection. The principles here may be quite different. While we want
the model for a given dataset to be as compressible as possible, perhaps we want to expose the model to data that makes the model as incompressible as
possible. The rationale for this prescription is that we want our models to learn to the fullest extent, with the hope that what
it is learning may be applied in a wide array of settings. This contention is at the heart of epiplexity.

We showed that epiplexity not only predicts better downstream generalization, but can be used to develop novel selection procedures that exceed state-of-the-art.
Perhaps surprisingly, we also showed that epiplexity has promise as a learning signal for generating purely synthetic data that improves out-of-distribution generalization.
The idea of synthesizing data for OOD generalization is entirely new, with many exciting future directions to explore.
Multi-modal and multi-objective generation, as well as efficiency
and scale, will be key considerations.
We note some limitations of our work. First, we conducted our experiments on small models, which are inherently constrained in their performance on downstream tasks, and additional experiments should be conducted on the scalability and generalizability of these methods across model architectures. In addition, in both data selection and generation, we relied on a tractable proxy for epiplexity rather than the quantity itself, and we leave to future work the theoretical guarantees that maximizing for this proxy during training maximizes the true epiplexity. Further ahead, we will in time have a better understanding of what abstract properties of data are shared across domains beyond epiplexity, so that we can generate increasingly universal data samples, leading to broadly improved generalization.

\textbf{Acknowledgements and Disclosure of Funding.} We thank Eric Elmoznino, Guillaume Lajoie, Anthony GX-Chen, and all members of the Wilson Lab for helpful feedback. This work was supported by DARPA AIQ HR00112590066, NSF CAREER IIS-2145492, NSF CDS\&E-MSS
2134216, Google’s TPU Research Cloud (TRC) program, and Lambda's Research Grant Program. ES is supported by the Meta AIM PhD Fellowship. SQ is supported by the Two Sigma PhD Fellowship. The authors declare no competing interests.

\bibliographystyle{unsrtnat}
\bibliography{references}

\newpage
\appendix

\paragraph{Appendix Outline} This appendix is organized as follows. In \cref{app:EpiCorrelation}, we analyze the correlation between epiplexity and downstream performance, contrasting it against the model's weight norm. In \cref{app:xDomain}, we provide the details of our cross-domain scaling laws, assess the accuracy of the fits, and present the ADO selector for comparison. In \cref{app:dataSelectEpi}, we derive the derivative of epiplexity gain with respect to the tokens seen from a domain. In \cref{app:epiDataSelectAlg} and \cref{app:epiDataGenAlg}, we give the full \texttt{EpiSelect} and \texttt{EpiGen} algorithms. In \cref{app:CPTokenBalance}, we describe the token-balancing procedure used to construct the Common Pile dataset. In \cref{app:dataSelectImplementation} and \cref{app:dataGenImplementation}, we provide the implementation details for the data selection and data generation experiments. In \cref{app:datagenTrainingDynamics}, we monitor the training dynamics of the epiplexity-guided generator. In \cref{app:evaluation}, we analyze the learner's language modeling quality across natural language distributions and study the effect of grounding the generator reward in the evaluation buffer. In \cref{app:syntheticDataEx}, we present truncated samples of the synthetic data produced by the generator.

\section{Epiplexity vs. Performance Analysis}\label{app:EpiCorrelation}
\begin{figure}[h]
    \centering
    \includegraphics[width=0.85\linewidth]{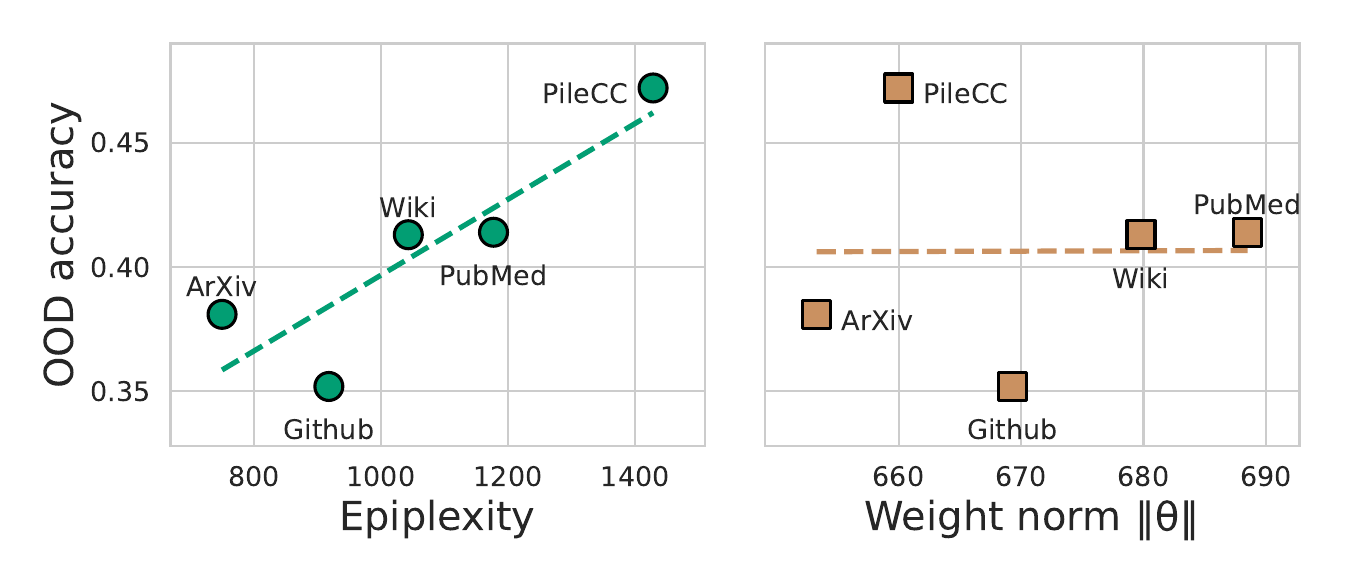}
    \caption{\textbf{Epiplexity tracks OOD downstream performance; weight norm does not.}
    Each point is a 124M model trained for 15B tokens on a single Pile domain, evaluated
    by mean zero-shot accuracy over seven held-out tasks. Epiplexity (left)
    is strongly correlated with accuracy (Pearson $r = 0.88$, $p = 0.05$; Spearman
    $\rho = 0.90$), whereas weight norm $\|\theta\|$ (right) is not ($r = 0.01$,
    $\rho = 0.20$). Dashed lines are least-squares fits, $n = 5$.}
    \label{fig:EpiCorrelation}
\end{figure}

To check that the association between epiplexity and downstream accuracy in
Section~\ref{sec:data_selection} reflects something specific to epiplexity, rather than
any incidental property of a trained checkpoint, we compare it against the model's weight norm $\lVert\theta\rVert$. Both quantities are scalar summaries of
a model trained to convergence on a single Pile domain, so if downstream accuracy
were merely a generic function of "how much" the model changed during training, we
would expect the weight norm to be similarly predictive.

Figure~\ref{fig:EpiCorrelation} shows this is not the case. Across the five domains,
epiplexity is strongly associated with OOD accuracy (Pearson $r = 0.88$, Spearman
$\rho = 0.90$). PileCC is highest on both, and ArXiv and GitHub are lowest for
both rankings. The weight norm, by contrast, shows no relationship with downstream
accuracy ($r = 0.01$, $\rho = 0.20$). With only five domains this is suggestive rather
than conclusive, but it indicates that the association is not explained by the
magnitude of weight change alone, and is consistent with epiplexity capturing something
about the data that generic checkpoint statistics do not.

\section{Scaling Laws} \label{app:xDomain}
\begin{figure}[h]
    \centering
    \begin{subfigure}{0.4\textwidth}
        \centering
        \includegraphics[width=\textwidth]{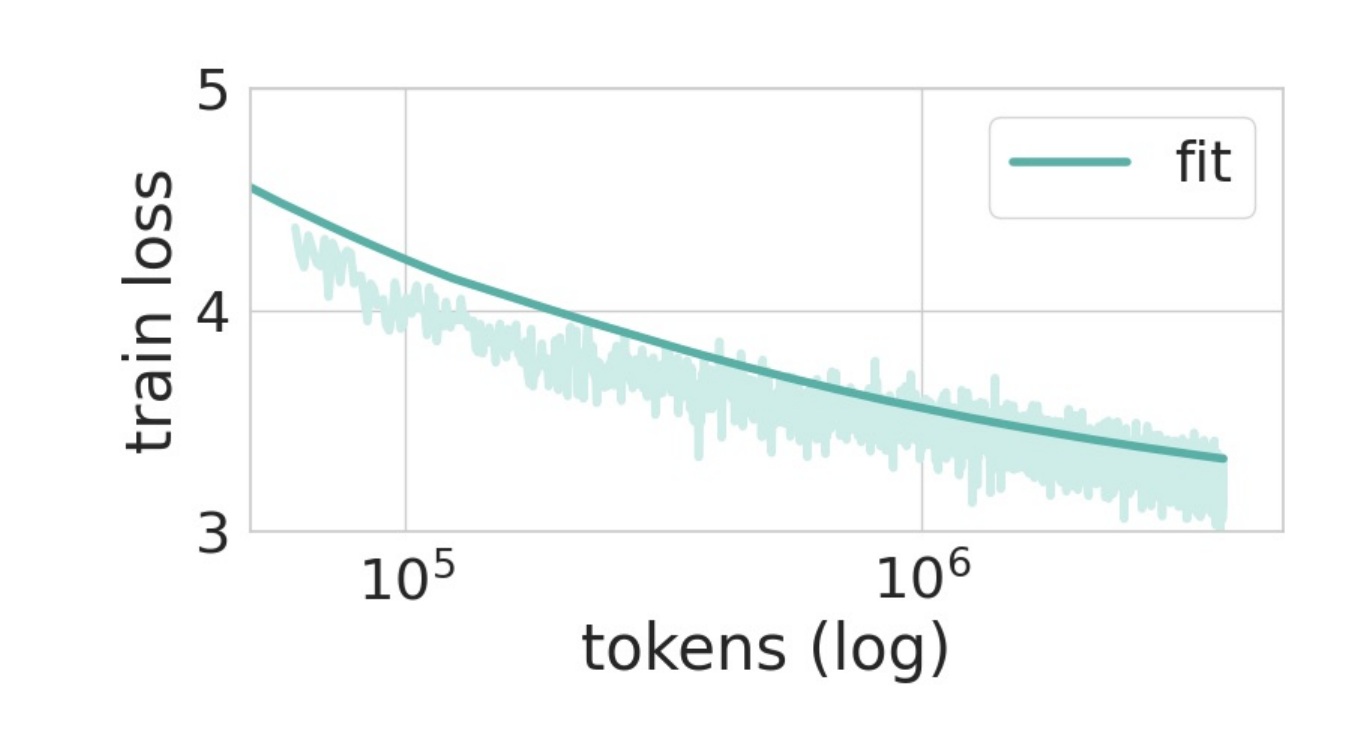}
    \end{subfigure}
    \hfill
    \begin{subfigure}{0.55\textwidth}
        \centering
        \includegraphics[width=\textwidth]{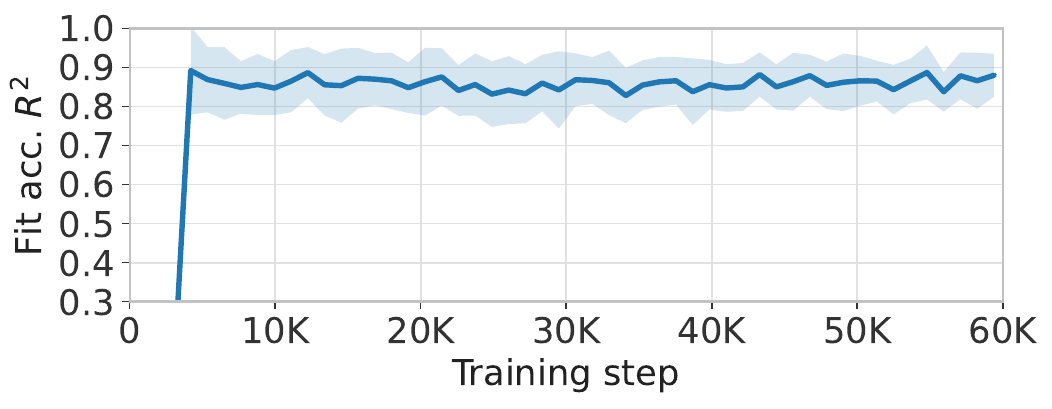}
    \end{subfigure}
    \caption{Panel (a): Example online cross-domain scaling-law fit for the
    \texttt{Common-Crawl} domain. Panel (b): Median fit accuracy over training over Common Pile domains, reported as $R^2$ and log-RMSE against the observed per-domain loss trajectories.}
    \label{fig:scaling_law_fits}
\end{figure}

Given some observed tokens $n_{k}^{(s)}$ and loss values $L_{k}^{(s)}$ for domains $k$ at different points in time $s=1, \dots, t$,
we fit a cross domain scaling law by solving the following optimization problem

$$\min_{\alpha, \beta, \epsilon, \gamma} \sum_{k,s} \text{HuberLoss}_{\delta}((L_{k}^{(s)})_{s=1}^{t}, (\hat{L}_{k}^{(s)}(\alpha, \beta, \epsilon, \gamma))_{s=1}^{t})
$$

where $\hat{L}_{k}^{(s)} = \hat{L}_{k}(n_{1}^{(s)}, \cdots, n_{K}^{(s)}) = \epsilon_{k} + \beta_{k} \left(\sum_{j=1}^{K} \gamma_{k,j} n_{j}^{(s)}\right)^{-\alpha_{k}}$.
We refit the scaling law every $1000$ iterations to make use of new available observed tokens and loss values.
Additionally, we optimize our parameters in the log-space, and work with $\alpha$, $\log \beta$, $\log \epsilon$ and $\log \gamma$. To satisfy the simplex constraints on $\gamma_{m,k}$, which are $\gamma_{m,k} \geq 0$ and $\sum_{k=1}^{K} \gamma_{m,k} = 1$, we apply a softmax transformation.

We initialize all parameters with a grid derived by the Cartesian product of the following ranges per parameter:
for $\alpha \in \{0.0, 0.1, \cdots, 0.8 \}$, $\log \beta \in \{-2.0, -1.0, \cdots, 6.0\}$, $\log \epsilon \in \{-2.0, -1.5, \cdots, 2.0\}$.
Finally, we sample $\gamma_{m,k} \sim \text{Dir}(p)$ via a Dirichlet distribution, where $p_m = 10$ and $p_k = 1$ if $k \neq m$.
In other words, the Dirichlet distribution concentrates more mass on the domain which is indicated by the row index. This creates an initialization grid of $729$ points which adds roughly $12$ sec of overhead per fit, as opposed to the 3 seconds \texttt{ADO} takes since it does not have to fit $\gamma_{m,k}$.
In practice, this increase is negligible as this overhead still constitutes less than $5\%$ of the training time for our $124$M model and less than $1\%$ for the $1.3B$ model. Panel (a) of Figure \ref{fig:scaling_law_fits} shows an example of our fitted scaling curve.

\subsection{Accuracy of the Scaling-Law Fits}
We assess how well the fitted scaling law reproduces the observed per-domain loss trajectories over the course of training for one of our runs. At every refit we compare the fitted function against the measured per-domain losses and report the coefficient of determination ($R^2$) and the root-mean-square error (RMSE) in log-loss space, matching the log-Huber objective. Panel (b) of Figure~\ref{fig:scaling_law_fits} shows that quality improves rapidly and then plateaus. $R^2$ rises to 0.9 within the first 4k steps and remains stable for the remainder of training. At the final step, the median $R^2$ across the 30 domains is 0.88 for the cross-domain law, with a median log-RMSE of 0.02 nats, meaning that predicted losses fall within roughly $2\%$ of the true observation values. Three domains achieved slightly lower $R^2$ (DM Mathematics, Enron Emails, and Ubuntu
IRC), which aligns with these domains all exhibiting near-flat, noisy loss curves, so their reduced $R^2$ does not indicate a problem with the fitting procedure. These results confirm that the fitted scaling laws are reliable for supporting domain selection decisions.

\subsection{ADO Selector for Comparison} \label{app:ADOSelector}
We provide the ADO selector for reference. ADO \citep{Jiang2024AdaptiveDO} fits a power scaling law of the form $\hat{L}(n;\alpha,\beta,\epsilon) = \epsilon + \beta n^{-\alpha}$ to each domain to inform the domain distribution $\pi$. Their preference distribution takes the form
\begin{align}
        \pi_{k} &\propto \mu_{k} \, \frac{\alpha_{k}}{n_{k}} (\hat{L}_{k}(n_{k}) - \epsilon_{k}) \, \lambda_{k}
\end{align}
which is notably different from our epiplexity-maximizing objective, as they do not model the effect of the different domains on one another and introduce a heuristic credit score term $\lambda_{k}$, whose role is to capture some interaction between the domains.

\section{Estimating Epiplexity Gain from Scaling Laws}\label{app:dataSelectEpi}
Below we derive the marginal epiplexity gain with respect to the number of tokens seen from domain $k$.
Although Equation~\ref{eq:dataSelectTotalEpi} writes the epiplexity sum over training steps $t$, we differentiate here with respect to the cumulative token count $n_k$, treating $S$ as a function of the token counts that the step index $t$ reaches. 
\begin{align*}
\frac{\partial \hat{S}_m(n_k)}{\partial n_k} &\approx \hat{S}_m(n_k + 1) - \hat{S}_m(n_k) \\
&= \sum_{i_k\leq n_k + 1} (\hat{L}_m(i_k) - \hat{L}_m(n_k + 1)) - \sum_{i_k\leq n_k} (\hat{L}_m(i_k) - \hat{L}_m(n_k)) \\
&= \left[ \sum_{i_k\leq n_k + 1} \hat{L}_m(i_k) - (n_k + 1) \hat{L}_m(n_k + 1) \right] - \left[ \sum_{i_k\leq n_k} \hat{L}_m(i_k) - (n_k) \hat{L}_m(n_k) \right] \\
&= \left[ -n_k\hat{L}_m(n_k + 1) \right] - \left[ - (n_k) \hat{L}_m(n_k) \right] \\
&= -n_k \Delta \hat{L}_m (n_k) \approx -n_k\frac{\partial \hat{L}_m}{\partial n_k}
\end{align*}

To arrive at Equation~\ref{eq:epiXDomainStepwise}, we sum over all $m$.

\section{Algorithms}
\subsection{EpiSelect}\label{app:epiDataSelectAlg}
Algorithm~\ref{alg:epiDataSelect} describes how we update the domain sampling
weights with \texttt{EpiSelect}. 

\begin{algorithm}
\caption{Epiplexity Guided Data Selection}
\label{alg:epiDataSelect}
\KwIn{Initial parameters $\theta$, scaling law $L$, scaling law parameters $\alpha, \beta, \gamma, \epsilon$, loss function $\mathcal{L}$, momentum coefficient $\omega \in (0,1)$, clipping parameter $\delta_{min}$, temperature $\tau > 0$, batch size $N \in \mathbb{N}$, and update frequency $\nu \in \mathbb{N}$.}
\KwOut{Trained parameters $\theta^{(T)}$}
\For{$t = 1, 2, \ldots, T$}{
\If{$\mod{(t, \nu)} = 0$}{
  $\alpha$, $\beta$, $\gamma$, $\epsilon$ = \texttt{FitScalingLaw} $((L^{(s)})_{s=1}^{t} , (n^{(s)})_{s=1}^{t})$ \\
  $\hat{S}(t) = \sum_{m = 1}^{K} \sum_{s=1}^{t} (\hat{L}_{m}^{(s)} - \hat{L}_{m}^{(t)})$ \tcp*{Estimate epiplexity}
  $\pi_{k} \propto \exp(\frac{1}{\tau} \frac{\partial \hat{S}(t)}{\partial n_{k}})$ \tcp*{Update distribution \cref{eq:piK}}
  $\pi_k \gets \rm{clip}(\omega \pi_{k} + (1 - \omega) \bar{\pi}_{k}, \delta_{min})$ \tcp*{Momentum}}
  $\mathcal{B} \sim \pi$, with $|\mathcal{B}| = N = \sum_{k} b_{k}^{(t)}$ \tcp*{Sample batch of data}
  $\bar{\pi}_{k} \leftarrow \frac{t}{t + 1} \bar{\pi}_{k} + \frac{1}{t+1} \pi_{k}$ \tcp*{Global mean}
  $L_{k}^{(t)} = \frac{1}{n_{k}^{(t)}} \sum_{x \in \mathcal{B}_{k}} \mathcal{L}(\theta^{(t)}, x)$ \tcp*{Compute per domain losses}
  $\theta^{(t+1)} = \texttt{OptimizerUpdate}(\theta^{(t)}, \nabla_{\theta} \mathcal{L}(\theta^{(t)}, \mathcal{B}))$
}
\end{algorithm}

\subsection{EpiGen}\label{app:epiDataGenAlg}

Algorithm \ref{alg:epiDataGen} describes the \texttt{EpiGen} training procedure.

\begin{algorithm}
\caption{Epiplexity Guided Data Generation}
\label{alg:epiDataGen}
\KwIn{Pretrained checkpoint $P_{\theta_0}$; learning rates $\eta_\ell, \eta_g$; generation temperature $\tau$; baseline EMA decay $\beta$; batch size $B$, buffer subsample size $M$; iterations $T$, learner steps per iteration $K$}
\KwOut{Generator $P_{\theta^{g}}$, learner $P_\theta$}
$\theta^{g} \leftarrow \theta_0, \theta \leftarrow \theta_0$ \tcp*{Init generator, learner}
$b \leftarrow 0, \mathcal{B} \leftarrow \emptyset$ \tcp*{Init baseline, buffer}
\For{$t = 1, 2, \ldots, T$}{
  $\mathcal{X}_t \leftarrow \{x_i\}_{i=1}^B \overset{\text{i.i.d.}}{\sim} P_{\theta^{g}}(\cdot \mid \tau)$ \tcp*{Generate batch}
    $\hat{\mathcal{B}}_{t-1} \sim \mathcal{B}_{t-1}, \quad    |\hat{\mathcal{B}}_{t-1}| = M$ \tcp*{Subsample buffer}
    $\hat{\mathcal{B}}_t \leftarrow \hat{\mathcal{B}}_{t-1} \cup \mathcal{X}_t$ \\
    $L_t^{pre} \leftarrow \frac{1}{M+B} \sum_{x \in \hat{\mathcal{B}}_t} \mathcal{L}(\theta_{t-1}, x)$ \tcp*{Loss before}
    $\theta_t \leftarrow \theta_{t-1}$ \\
    \For {$k = 1, 2, \ldots, K$}{
    $\theta_{t} \leftarrow \theta_t - \frac{\eta_\ell}{B} \sum_{x \in \mathcal{X}_t} \nabla_{\theta_t} \mathcal{L}(\theta_t, x)$ \tcp*{Learner update}
    }
    $L_{t}^{post} \leftarrow \frac{1}{M+B} \sum_{x \in \hat{\mathcal{B}}_t} \mathcal{L}(\theta_{t}, x)$ \tcp*{Loss after}
    $r_t \leftarrow L_t^{pre} - L_{t}^{post}$ \\
    $\theta^{g} \leftarrow \theta^{g} + \frac{\eta_g}{B} [ (r_t - b) \cdot \sum_{x \in \mathcal{X}_t} \nabla_{\theta^{g}} \log P_{\theta^{g}}(x)]$ \tcp*{Generator update}
    $b \leftarrow \beta \cdot b + (1 - \beta) \cdot r_t$ \\
    $\mathcal{B}_t \leftarrow \mathcal{B}_{t-1} \cup \mathcal{X}_t$ \\
}
\end{algorithm}

\section{Common Pile Token Balancing} \label{app:CPTokenBalance}
For the Common Pile dataset, we streamed up to 2B tokens for each of the 6 out of 30 largest domains and the maximum token count for the 24 out of 30 smaller domains, which totaled 29.7B tokens. Our data selection training requires 15.7B tokens over the 60,000 training steps. Thus, this construction serves as the token-balanced baseline, as truly balancing the token counts per domain would have to match the single-millions of tokens in the smallest domains, which would not provide sufficient unique tokens for training. Our \texttt{Natural} baseline, which selects proportionally to the default domain sizes, reported in Table~\ref{tab:downstream-std} was run on this construction.

\section{Implementation}

\subsection{Data Selection} \label{app:dataSelectImplementation}

\paragraph{\textbf{Models}} Our experiments use a decoder-only transformer from the LLaMA 2 family in two model parameter sizes 124M and 1.3B.
The 124M model has 12 layers, 12 attention heads and an embedding size of 768.
The 1.3B model has 24 layers, 16 attention heads and an
embedding size of 2048. Both models use SwiGLU feed-forward layers,
RMSNorm, and rotary position embeddings (RoPE).
The vocabulary follows the GPT-NeoX-20B tokenizer (50,304 tokens) and the context length is 1,024 tokens.
Parameters are stored in float32. Forward passes run in bfloat16.
Models are implemented in JAX with Equinox for module definitions and use Optax for optimization.

\paragraph{\textbf{Training}} Each model is trained for 60,000 steps with a global batch size of 256 sequences (262,144 tokens per step, ~15.7B tokens total).
We use AdamW with a peak learning rate of $10^{-3}$, $\beta_2$ = 0.95, weight decay $10^{-4}$, a 500-step linear warm-up, and cosine decay to $10^{-5}$.
We update the data-selection weights every 1,000 steps, after an initial 1,000 steps on the natural (by domain size) distribution to accumulate sufficient per-domain loss history for scaling-law fitting.

\paragraph{\textbf{Compute}} All of our training was done on a single Cloud TPU v4-32 slice (4 host VMs, each with 8 TPU v4 chips).
When launched, the model is replicated across all 32 chips, and each chip processes 8 sequences per step.
Wall-clock training time is approximately 8 hours per 60,000-step run
for the 124M model and 29 hours for the 1.3B model.

\subsection{Data Generation} \label{app:dataGenImplementation}
\paragraph{\textbf{Models}} We initialize both the generator and learner models with pretrained GPT-2 weights (117M parameters, 12 layers, 12 attention heads, 768-dimensional embeddings), loaded via HuggingFace Transformers. The frozen reference model used for computing pretrained perplexity is a copy of the same model with gradients disabled. All models use the GPT-2 tokenizer (50,257 tokens). Models are implemented in PyTorch with gradient check-pointing enabled during training.

\paragraph{\textbf{Training}} We use a REINFORCE-based training loop with $T = 6{,}000$ generator steps and $K = 10$ learner gradient steps per generator step, for a total of 60,000 learner updates per run. At each step the generator produces a batch of 32 sequences of 512 tokens at temperature $\tau = 1.0$. Each generated batch is appended to the accumulating buffer, which grows during training. At each step, 16 batches are sampled uniformly from the buffer and combined with the current batch to evaluate the learner. The epiplexity-based reward is computed as the decrease in learner loss before and after training on the evaluation batch. We implement gradient norm clipping (1.0) to reduce variance during the REINFORCE update. In the EMA baseline case, we use a $\beta=0.99$. The learner uses AdamW with lr $= 10^{-5}$ and weight decay $10^{-2}$, and the generator uses AdamW with lr $= 10^{-6}$ and weight decay $10^{-2}$. Finally, we compute the validation loss every 1000 learner steps on 500K tokens of OpenWebText \citep{Gokaslan2019OpenWeb}, which is in the original training data of GPT-2, to measure drift from the original pretraining distribution.

\paragraph{\textbf{GLUE fine-tuning}} For each learner, we fine-tune all weights along with a linear classification head on the seven GLUE tasks (CoLA, SST-2, MRPC, QQP, MNLI, QNLI, RTE) for 3 epochs using a learning rate of $2 \times 10^{-5}$ and batch size 32, then evaluate the model on the development set.

\paragraph{\textbf{Compute}} We conduct all runs on 4 NVIDIA L40S GPUs with PyTorch DDP. Wall-clock training time is approximately 10 hours per 60,000-step run.

\section{Monitoring Data Generation Training Dynamics} \label{app:datagenTrainingDynamics}

\begin{figure}
    \centering
    \includegraphics[width=\linewidth]{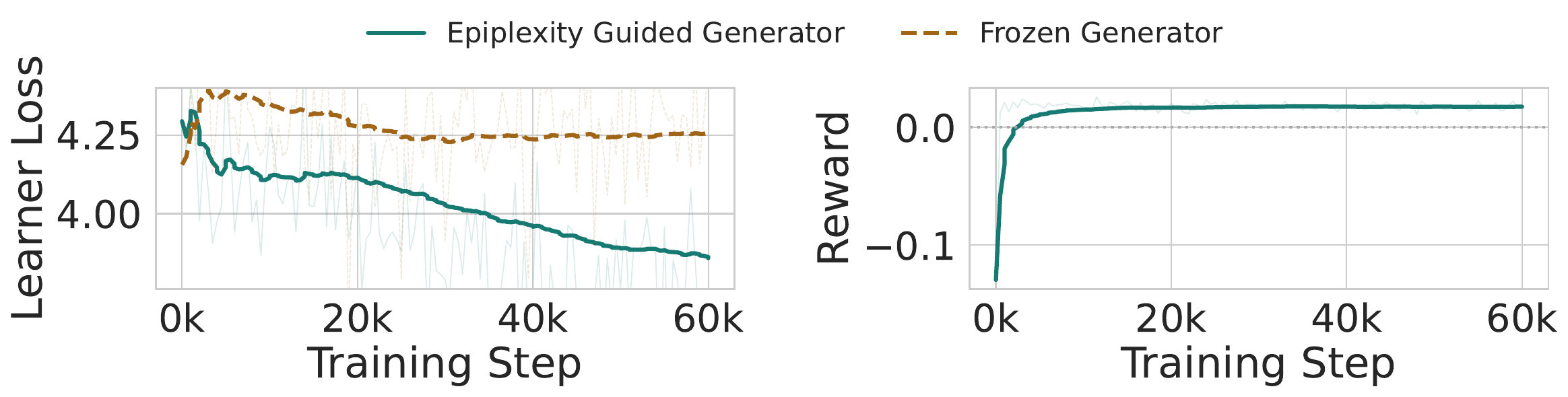}
    \caption{\textbf{Training dynamics of the epiplexity-guided generator.} (\textit{Left}) Learner loss on generated data decreases over training for the epiplexity-guided generator, while remaining flat for the frozen generator baseline. (\textit{Right}) REINFORCE reward increases over training, indicating that the generator learns to produce informative synthetic data throughout. No reward is computed for the frozen generator. }
    \label{fig:datagenTraining}
\end{figure}

Figure~\ref{fig:datagenTraining} shows the training dynamics (learner train loss and reward) to verify that the REINFORCE training loop functions as intended. The learner loss is measured on the current batch of generated data after $K$ learner gradient steps. Both runs initialize from the same pretrained checkpoint and therefore start at similar losses. Over training, the epiplexity-guided generator produces data on which the learner achieves progressively lower loss, indicating that the generator learns to synthesize data more aligned with the learner's current state. The frozen generator, by contrast, produces a stationary data distribution and the learner loss remains relatively flat throughout. Meanwhile, the reward is the epiplexity signal used to update the generator via REINFORCE. The reward rises steadily over training and remains positive, confirming that the generator consistently produces data that improves the learner beyond its EMA baseline. The frozen generator is never updated, so we do not compute the reward.

\section{Evaluation} \label{app:evaluation}
\subsection{Held-Out Perplexity Benchmarks}
  \label{app:oodBenchmarks}

  Table~\ref{tab:epigenTAB1} reports the perplexity of each learner on held-out natural text from three
  corpora. OpenWebText \citep{Gokaslan2019OpenWeb} is an open-source reproduction of GPT-2's pretraining
  corpus and serves as the in-distribution benchmark. SlimPajama \citep{soboleva2023slimpajama} and
  FineWeb \citep{penedo2024fineweb} are web-scale corpora that GPT-2 was not trained on and serve as the
  out-of-distribution benchmarks. Since \texttt{EpiGen} and its ablations train the learner only on
  synthetic text, none of the three corpora is seen during training.

  \paragraph{Evaluation sets.} For each corpus we stream documents in their released order, tokenize them
  with the GPT-2 BPE tokenizer, concatenate the token streams, and keep the first 5M tokens. We use the
  \texttt{train} split of OpenWebText, the \texttt{validation} split of SlimPajama, and the
  \texttt{sample-10BT} subset of FineWeb. The same fixed 5M-token slices are used for every model, so all
  rows of Table~\ref{tab:epigenTAB1} are scored on identical tokens.

  \paragraph{Perplexity.} Each 5M-token stream is split into non-overlapping windows of 513 tokens
  (a 512-token context and its next-token targets), giving roughly $9.7$k windows per corpus. We compute
  the mean next-token cross-entropy $\mathcal{L}$ in nats over all windows, in batches of 32 with no
  document boundaries or attention masking, and report $\text{PPL} = \exp(\mathcal{L})$. 
  The Pretrained row scores the GPT-2 checkpoint that initializes every learner, and each remaining row
  scores the final learner of a training run. Learner rows are mean $\pm$ sample standard deviation over
  three training seeds; the Pretrained row is a single deterministic evaluation.

  \subsection{Generator Sample Statistics}
  \label{app:generatorStats}

  The right-hand columns of Table~\ref{tab:epigenTAB1} characterize the synthetic text that each
  generator produces at the end of training. For every generator we draw fresh samples exactly as during
  training: ancestral sampling at temperature $\tau = 1$ from the \texttt{<|endoftext|>} prompt, with no
  top-$k$ or nucleus truncation, for 512 tokens per sequence. We draw 8 batches of 32 sequences
  (256 sequences, roughly $131$k tokens) with a fixed sampling seed. The \texttt{FrozenGen} generator is
  never updated, so its statistics are those of pretrained GPT-2 and its standard deviation reflects
  sampling variance alone; all other rows are mean $\pm$ sample standard deviation over the three seeds' generators.
  The \textcolor{gray}{Real text} row applies the same statistics to 256 held-out OpenWebText sequences of
  512 tokens, disjoint from the evaluation sets of Appendix~\ref{app:oodBenchmarks}.

  \paragraph{GPT-2 perplexity (naturalness).} The perplexity that the \emph{pretrained} GPT-2 checkpoint
  assigns to the samples, $\exp$ of the mean next-token cross-entropy over all sampled tokens. Text that
  resembles GPT-2's training distribution scores close to the real-text value of $25.4$, whereas text that is
  degenerate (highly repetitive) scores far below it and text that approaches noise scores far above it.

  \paragraph{Entropy (nats).} The average entropy of the generator's own next-token distribution,
  $\mathbb{E}_t\big[-\sum_v p_\theta(v \mid x_{<t}) \log p_\theta(v \mid x_{<t})\big]$, evaluated
  teacher-forced on its own samples and averaged over all positions and sequences. It measures how
  spread out the generator's sampling distribution is. A low value indicates a near-deterministic
  generator, a high value one that samples close to uniformly over the vocabulary. Entropy is a property
  of the generator, so it is undefined for the real-text row.

  \paragraph{4-gram repetition.} The fraction of 4-gram positions within a sequence whose 4-gram has
  already occurred earlier in the same sequence, pooled over all sampled sequences~\citep{welleck2020unlikelihood}. A value of $0$ means no 4-gram is
  ever repeated within a sequence, and a value near $1$ means the sequence loops. Real OpenWebText text
  has a small nonzero rate ($0.034$) from naturally recurring phrases.

\subsection{Reward Grounding in the Evaluation Buffer} \label{app:OWTBuffer}
\begin{figure}[h!]
    \centering
    \includegraphics[width=\linewidth]{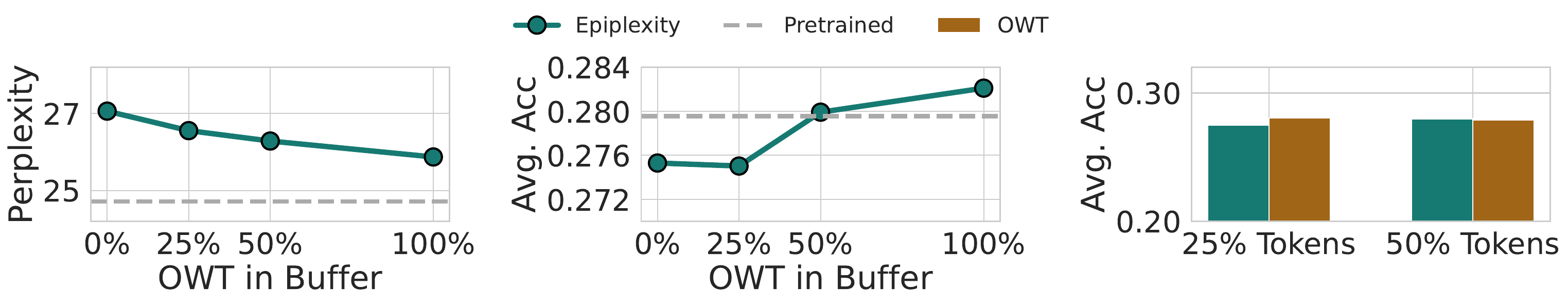}
    \vspace{-5mm}
    \caption{
        \textbf{Mixing OWT data into the evaluation buffer grounds the generator reward.}
        \textit{(Left)} The learner's perplexity decreases monotonically as the fraction of OWT data in the buffer increases from 0\% to 100\%.
        \textit{(Center)} Average zero-shot accuracy across 10 LM Eval Harness tasks similarly improves with more real data in the buffer.
        \textit{(Right)} At matched token budgets, learners trained via epiplexity-guided synthetic data perform comparably to models fine-tuned directly on the same number of real OWT tokens.
    }
    \label{fig:OWTAblation}
    \vspace{-3mm}
\end{figure}

While epiplexity-guided training consistently improves downstream performance in fine-tuning based tasks, it also subjects the learner to data which diverges from
natural language and leads to an increase in validation perplexity on web text.
In this section, we intervene on the training algorithm by mixing fractions of OWT data into the evaluation buffer \citep{Gokaslan2019OpenWeb}.
In the ideal case, we would compute the change in epiplexity as the difference in learner loss over all previous data (pretraining and synthetic) that the learner has
been trained on following Equation \ref{eq:prequential}.
Thus, as pretraining data greatly outsizes the synthetic data, we approximate this setting by sampling from the OWT dataset when constructing our evaluation buffer to
compute the reward.

Additionally, we hypothesize that the inclusion of batches of OWT data grounds the generator's predictions in its initial distribution, which reduces the effect of potential self-referential rewards. Concretely, we hold all hyperparameters fixed and ablate the proportion of OWT data in the buffer between 0\%, 25\%, 50\%, and 100\%.
We evaluate each learner model on OWT perplexity as well as on the same suite of 10 zero-shot tasks from LM Evaluation Harness as in prior experiments (ARC, BBQ, BoolQ, CSQA, HellaSwag (HSwag), LAMBADA, OBQA, PIQA, SciQ, and WinoGrande) \citep{eval-harness}.
Figure \ref{fig:OWTAblation} shows that increasing the fraction of OWT data in the buffer reduces OWT perplexity toward the pretrained GPT-2 baseline and improves average zero-shot accuracy on LM Eval Harness tasks beyond the pretrained GPT-2 baseline.
Although the learner is never directly trained on real data, it is reasonable to assume being indirectly exposed to OWT data in the buffer
might be driving the improvement.
To measure this effect, we compared our learner to token-matched baselines where we actually train the learner on OWT data and not synthetic generations.
Surprisingly, we find that the learner trained via epiplexity-maximizing synthetic data performs comparably to the corresponding baselines, which
not only validates our procedure but also shows that the synthetic data from the generator is as informative as real natural web text.

\section{Synthetic Data Outputs} \label{app:syntheticDataEx}
We examined the samples of synthetic data generated by the epiplexity-trained generator at step 1,000 and step 60,000 (the final step) of training and show truncated snippets of text in Figure \ref{fig:samples}. Notably, the generator receives no verifier signal and is \textit{never} exposed to real data during either training or in the evaluation buffer in this case. The epiplexity reward is computed entirely from the learner's loss decrease over a subsample of the previous synthetic data samples. Despite this, the generator maintains consistent web-text style throughout training, producing news articles, Q\&As, and blog-style content that are stylistically coherent with the OpenWebText pretraining distribution. The qualitative similarity between early and late samples suggests that the generator does not degrade or collapse under the purely self-referential reward, but instead sustains a stable generative distribution across the full training run.
\clearpage
\begin{figure*}[h!]
  \footnotesize
  \begin{minipage}[t]{0.48\linewidth}
  \textbf{Step 1{,}000}
  \medskip\hrule\medskip
  Maybe the bigger problem lies with this form of advertising. And perhaps I've answered everything. Save these emails and a free copy of my book instead, but
  just in case. Editor's note: This does not apply to tips and corrections. All parts of this article are intended for educational and informational purposes
  only, and do not necessarily reflect the views or opinions of TechCrunch and its parent company, Storage Spaces, Inc. Content Gaining Effectes. (a) The
   polar opposite. That's what happens when we assume that one form of style/coloring we use will boost effectiveness. [\ldots]
  \medskip\hrule\medskip
  The Grand Prix Jaipur is to commence this spring. What's more, the P1 class's roster includes Vandoorne, Bora Senna, Jorge Renato, Robbie Hulkenberg, Benetton
  Nikola Karasevic, Caleb Ewan, Angelica Hutschi, Travis Meyer, Benjamin Marko, Frank Wolff and Masaaki Matsumoto. The 2016 Series champion consists of the result
   of the hit Midsport Dragonslaying 20m. Bora, Senna and Jamie Stewart returned to Hamilton for Barber Motorsports' quadrupleheader series four years ago.
  [\ldots]
  \end{minipage}
  \hfill
  \begin{minipage}[t]{0.48\linewidth}
  \textbf{Step 60{,}000 (Final)}
  \medskip\hrule\medskip
  Europe is ``divesting itself of its remittances from Russia,'' Chancellor George Osborne told Parliament on Monday, the second such threat from the president of
   Russia since the meeting with UN delegates last December. The decision has alarmed European officials worried that Putin's economic dumping of assets to Russia
   is hurting thousands of EU citizens and their businesses in the troubled former Soviet bloc when Russia is perceived as something of a chess-playing player.
  [\ldots]
  \medskip\hrule\medskip
  Do I write a 3-D magazine or not? No. This depends on the style, strength and variety of stories. Publishing and website design are done by hand or by skilled
  3-D artists that are expert in computer animation at GE. Oh, so there are of course restrictions on what people can produce in 3 direct print magazines like
  these ``Fairyballs'' and can only print a wide range of images, so ``99\% 2002 images'' variety? True, you can print out anything from red photographs to
  precious candied bunches of paper alongside your old 200 page collectibles!
  [\ldots]    \end{minipage}
  \caption{\textbf{Truncated samples of synthetic data generated at step 1{,}000 (left) and step 60{,}000 (right).} Qualitatively, the generator produces web-text-like content throughout training (e.g. news articles, Q\&As, blog posts), consistent with its OpenWebText pretraining distribution. While the content is stylistically plausible, it may be factually hallucinated.}
  \label{fig:samples}
  \end{figure*}

\end{document}